\documentclass[twocolumn]{autart}    

\usepackage{cite}
\usepackage{amsmath,amssymb,amsfonts}
\usepackage{graphicx}
\usepackage{textcomp}
\usepackage{xcolor}
\usepackage{diagbox}
\usepackage{graphicx} 
\usepackage{epstopdf}
\usepackage{algorithmic}
\usepackage{algorithm}

\usepackage{pdfpages}
\usepackage{subfigure}
\usepackage{tabularx}
\usepackage{booktabs}
\usepackage{multirow}

\makeatletter
\newcommand{\removelatexerror}{\let\@latex@error\@gobble}
\makeatother

\begin{document}

\begin{frontmatter}

\title{Dealing with Collinearity in Large-Scale\\ Linear System Identification Using Gaussian Regression} 

\thanks[footnoteinfo]{This paper was not presented at any IFAC
meeting. Corresponding author W.~Cao.}

\author[wnq]{Wenqi Cao}\ead{wenqicao@sjtu.edu.cn},    
\author[giapi]{Gianluigi Pillonetto}\ead{giapi@dei.unipd.it},               

\address[wnq]{Department of Automation, Shanghai Jiao Tong University, Shanghai, China}  
\address[giapi]{Department of Information Engineering, University of Padova, Padova, Italy}             


\begin{abstract}                          
Many problems arising in control
require the determination of a mathematical model of the application.
This has often to be performed starting from input-output data,
leading to a task known as system identification in the engineering literature.
One emerging topic in this field is estimation of networks consisting of several interconnected
dynamic systems.
We consider the linear setting assuming that system outputs are the result of many correlated inputs,
hence making system identification severely ill-conditioned.
This is a scenario often encountered when modeling complex systems composed by many sub-units with feedback
and algebraic loops.
We develop a strategy cast in a Bayesian regularization framework
where any impulse response is seen as realization of a zero-mean Gaussian process.
Any covariance is defined by the so called stable spline kernel which includes
information on smooth exponential decay.
We design a novel Markov chain Monte Carlo scheme able to reconstruct the
impulse responses posterior by efficiently dealing with collinearity.
Our scheme relies on a variation of the Gibbs sampling
technique: beyond considering blocks forming a partition of the parameter space,
some other (overlapping) blocks are
also updated on the basis of the level of collinearity of the
system inputs. Theoretical properties of the algorithm are
studied obtaining its convergence rate. Numerical experiments
are included using systems containing hundreds of impulse
responses and highly correlated inputs.
\end{abstract}

\end{frontmatter}

\section{Introduction}
Large-scale dynamic systems arise in many scientific fields
like engineering, biomedicine and
neuroscience, examples being sensor networks, power grids and the brain \cite{Hagmann2008,Hickman2017,Pagani2013,Prando2020}.
Modeling these complex physical systems, starting from input-output data,
is a problem known as system identification in the literature  \cite{Ljung:99,Soderstrom}.
It is a key preliminary step
for prediction and control purposes \cite{Ljung:99,Soderstrom}.
Often, the systems under study can be seen as networks composed of
a large set of interconnected sub-units.
They can thus be described through many nodes which can communicate each other
through modules driven by measurable inputs or noises \cite{Chiuso2012,Materassi2010,VdH2013,Yue2021,CPL21, CLP21}.

In this paper we assume that linear dynamics underly our large-scale dynamic system
and focus 
on the identification of Multiple Inputs Single Output (MISO) stable models.
The problem thus reduces to estimation of impulse responses. 
In addition, the system output measurements are assumed to be the result of many measurable and
highly correlated inputs, possibly also low-pass with poor excitation of system dynamics. 
Such scenario is not unusual in dynamic networks
since modules interconnections give rise to feedback
and algebraic loops \cite{Bazanella2017,Goncalves2008,Hendrickx2019,Weerts2018,Fonken2020,Ramaswamy2021b}. 
This complicates considerably the identification process:
low pass and almost collinear inputs lead to severe ill-conditioning \cite[Chapter 3]{CollBook}\cite{ChiusoPicci04},
as described also in real applications like \cite{LLB16,MPJR10,PV97}.
This makes especially difficult the use of the classical approach to system identification \cite{Ljung:99,Soderstrom}, where
different parametric structures have to be postulated. For instance, the introduction of rational transfer functions
to describe each impulse response leads to high-dimensional nonconvex optimization problems.
In addition, the system dimension (i.e. the discrete order of the polynomials defining the rational transfer functions) has typically to be estimated from data,
e.g. using AIC or cross validation \cite{Akaike:74,Hastie01}.
This further complicates the problem
due to the combinatorial nature of model order selection.

The problem of collinearity in linear regression has been discussed for several decades \cite{CollBook,Srisa21,Zhang18}.
Two types of methods are mainly considered. One consists of eliminating some related regressors, the other of exploiting
special regression methods like LASSO, ridge regression or partial least squares \cite{Srisa21}.
The main idea underlying these approaches is to eliminate or decrease the influence of correlated variables.
This strategy can be useful when the goal is to obtain a model useful for predicting future data
from inputs with statistics very
similar to those in the training set.

This paper follows a different route based on
Bayesian identification of dynamic systems via Gaussian regression \cite{Rasmussen,Bell2004,SurveyKBsysid,BottegalQuant2017}.
It exploits recent intersections between
system identification and machine learning which involve Gaussian regression,
see also \cite{Wei,GP2023} for other lines which combine dynamic systems and deep networks.
In particular, we couple the stable spline prior
proposed in \cite{SS2010,PillACC2010,SS2011} 
with a Markov chain Monte Carlo (MCMC) strategy \cite{Gilks} suited to overcome collinearity.
These two key ingredients (stable spline and MCMC) of our algorithm are now briefly introduced.
The stable spline kernel is a nonparametric description of stable systems: it embeds
BIBO stability, one of the most important concepts in control which ensures that
bounded inputs always produce bounded outputs.
It has shown some important advantages in comparison
with classical parametric prediction error methods \cite{Ljung:99,SurveyKBsysid,PillonettoBook2022}.
It models an impulse response as a zero-mean Gaussian process whose covariance includes information on
smooth exponential decay and depends on two hyperparameters:
a scale factor and a decay rate \cite{PillonettoTAC2021,PillonettoMLrob2015,ARAVKIN2012}.
The main novelty in this paper is that the stable spline kernel is implemented
using a Full Bayes approach (hyperparameters are also modeled as random variables)
coupled with an MCMC strategy able to deal with collinearity.
MCMC is a class of algorithms that builds a suitable Markov chain
which converges to the desired target distribution. In our case,
the distribution of interest is the joint posterior of hyperparameters and impulse responses.
To deal with collinearity, we design a new formulation of a particular MCMC scheme known as Gibbs sampling
\cite{Gilks}
able to visit more frequently those parts of the parameter space 
mostly correlated. In particular,  beyond considering blocks forming a partition of the parameter space,
some other (overlapping) blocks are also updated
on the basis of the level of inputs correlation.
Theoretical properties of this algorithm are studied obtaining its convergence rate.
Such analysis is then complemented with numerical experiments. This allows to quantify the advantages
of the new identification procedure both under a theoretical and an empirical framework.

The structure of the paper is as follows. Section~\ref{secProb} reports the problem statement by introducing the measurement model and the prior model on the unknown impulse responses. In Section~\ref{secAlg}, we describe the system identification procedure, discuss the issues regarding the selection of the overlapping blocks to be updated during the MCMC simulation and the impact of stable spline scale factors on the estimation performance.
Convergence properties of our algorithm are illustrated in Section~\ref{secConvg}.
Section~\ref{secExp} reports three numerical examples which highlight accuracy and efficiency of
the identification procedure under collinearity. Conclusions then end the paper while Appendix
reports the proof of the main convergence result here obtained. 

\section{Problem Formulation and Bayesian model}\label{secProb}

\subsection{Measurements model}
Consider a node in a dynamic network associated
to a measurable and noisy output denoted by $y$.
We assume that such output is the result of many inputs $u_k$
which communicate with such node through some linear and stable dynamic systems.
The transfer function of each module is assumed to be rational, hence
the $z$-transform of any impulse response is the ratio of two polynomials.
This is one of the most important descriptions of dynamic systems encountered in nature.
Specifically, the measurements model is
\begin{equation}\label{MM1}
y(i) =  \sum_{k=1}^m F_k u_k(i) + e(i), \quad i=1,\ldots,n
\end{equation}
where the $F_k$ are stable rational transfer functions sharing a common denominator,
$y(i)$ is the output measured at $t_i$, $u_k(i) $ is the known input entering the $k$-th channel at time $t_i$,
$m\geq 2$ is the number of modules influencing the node. 
Finally, the random variables $e{(i)} $ form a white Gaussian noise
of variance $\sigma^2$. The problem is to estimate the  $m$
transfer functions from the input-output samples. The difficulty we want to face is that
the number of unknown parameters, given by the coefficients of the polynomials entering the $F_k$,
can be relatively large w.r.t. the data set size $n$ 
and some of the $u_k$ may be highly collinear.
Thinking of the inputs as realizations of stochastic process,
this means that the level of correlation among the $u_k$ can be close to 1.
Also, the model order, defined by the degrees of the polynomials in the $F_k$,
can be unknown.

\subsection{Convexification of the problem and Bayesian framework}

An important convexification of the model \eqref{MM1} is obtained
approximating each stable $F_k$ by a FIR model of  order $p$ sufficiently large.
This permits to rewrite
\eqref{MM1} in matrix-vector form.
In particular, if $G_k \in \mathbb{R}^{n \times p}$ are suitable Toeplitz matrices containing
past input values, one has 
\begin{equation}\label{MM2}
Y = \left( \sum_{k=1}^m G_k \theta_k \right) + E = G \theta + E
\end{equation}
where the column vectors $Y,\theta_k$ and $E$ contain, respectively,
the $n$ output measurements, the $p$ impulse response coefficients defining the
$k$-th impulse response and the $n$ i.i.d.
Gaussian noises. Finally, on the rhs $G=[G_1 \ldots G_m]$
while $\theta$ is the
column vector which contains all the impulse response coefficients contained in the $\theta_k$.

A drawback of the formulation \eqref{MM2} is that, even if FIR models
are easy to estimate through linear least squares, they may suffer of large variance.
This problem is especially relevant in light of the assumed collinearities
among the inputs: the regression matrix $G$
turns out to be ill-conditioned, with a very large condition number, so that small output noises
can produce large estimation errors.
This problem is faced by the Bayesian  linear system identification approach documented in
\cite{SS2010,SurveyKBsysid}: for known covariance, the $\theta_k$ are modeled as independent zero-mean Gaussian vectors
with covariances proportional to the stable spline matrix  $K \in \mathbb{R}^{p \times p}$.
Such matrix encodes smooth exponential decay information on the coefficients of $\theta_k$:
the $i,j$-entry of $K$ is
\begin{equation}\label{SS}
K(i,j) = \alpha^{\max(i,j)}. 
\end{equation}
where $\alpha$ regulates how fast the impulse response variance decays to zero.
This parameter is assumed to be known in what follows
to simplify the exposition.

Differently from \cite{SS2010}, the covariances of the $\theta_k$ depend on scale factors $\lambda_k$
which are seen as independent random variables.
Hence, beyond the impulse responses conditional on the $\lambda_k$,
also the stable spline covariances $\lambda_k K$ become stochastic objects.
The $\lambda_k$ follow an improper Jeffrey's distribution
\cite{Jeffreys1946}. Such prior in practice includes only nonnegativity information and is defined by
\begin{equation}\label{Hp1}
p(\lambda_k) \sim \frac{1}{\lambda_k}
\end{equation}
where here, and in what follows, $p(\cdot)$ denotes a probability density function.
 In its more general form, our model assumes that all the scale factors $\lambda_k$ are mutually independent.
But, later on, we will also see that constraining all of them to be the same can be crucial to deal with collinearity.
The noise variance $\sigma^2$ is also a random variable (independent of $\lambda_k$) and follows the Jeffrey's prior.
Finally, for known $\lambda_k$ and $\sigma^2$, all the $\theta_k$ and the noises in $E$ are assumed mutually independent.

\section{Bayesian regularization: enhanced Gibbs Sampling using overlapping blocks}\label{secAlg}

\subsection{Bayesian regularization}

Having set the Bayesian model for system identification, our target is now
to reconstruct in sampled form the posterior of impulse responses and hyperparameters.
This then permits to compute minimum variance estimates of the system coefficients
and uncertainty bounds around them. To obtain this, the section is so structured.
First, we show that this objective could be obtained resorting to Gibbs sampling,
dividing the parameter space in (non overlapping) blocks which are sequentially updated.
Since we assume that the number of impulse responses can be considerable,
the dimension of the blocks cannot be too large. For instance, due to computational reasons,
it would be impossible to update simultaneously
all the components of $\theta$. One solution is to resort to a classical Gibbs sampling scheme
where any impulse response $\theta_k$ is associated with a single block. However, we will see that, in presence
of high inputs collinearity, this approach does not work in practice: slow mixing affects the generated Markov chain.
The last parts of this section overcome this problem by designing a new
stochastic simulation scheme. It exploits additional overlapping blocks
to improve the MCMC convergence rate. 

\subsection{Gibbs sampling}
Let $\mathcal{N}(\mu,\Sigma)$ denote the multivariate Gaussian density of mean $\mu$ and covariance matrix $\Sigma$.
From the model assumptions reported in Section~\ref{secProb}, one immediately obtains the likelihood of the data
and the conditional priors of impulse responses and noises as
\begin{subequations}\label{priDis}
\begin{equation}
    Y \mid (\{\theta_k\},\{\lambda_k\},\sigma^2)\sim  \mathcal{N}(G\theta, \sigma^2I),
\end{equation}
\begin{equation}\label{priDisTheta}
    \theta_k \mid \lambda_k \sim \mathcal{N}(0,\lambda_k K), 
\end{equation}
\begin{equation}
    E \mid \sigma^2 \sim \mathcal{N}(0,\sigma^2 I). 
\end{equation}
\end{subequations}

Using Bayes rule, standard calculations lead also to the following
\emph{full conditional distributions} of the variables we want to estimate: 
\begin{subequations}\label{postDis}
\begin{equation}\label{postDisLambdak}
    \lambda_k \mid (Y, \sigma^2, \{\theta_j\}, \{\lambda_j\}_{j\neq k}) \sim {\mathcal{I}_g}(\frac{p+1}{2},~ \frac{1}{2}\theta_k'K^{-1}\theta_k),
\end{equation}
\begin{equation}
    \sigma^2 \mid (Y,\{\theta_j\}, \{\lambda_j\})\sim  {\mathcal{I}_g} (\frac{n}{2}, \frac{1}{2}{\Vert Y-G\theta\Vert^2}),
\end{equation}
\begin{equation}\label{theta_k}
    \theta_k \mid (Y,\sigma^2, \{\lambda_j\}, \{\theta_j\}_{j\neq k}) \sim \mathcal{N}(\hat{\mu}_k, \hat{\Sigma}_k),
\end{equation}
where ${\mathcal{I}_g(\cdot,\cdot)}$ denotes the inverse Gamma distribution and
\begin{equation}\label{muk}
    \hat{\mu}_k= \hat{\Sigma}_k\frac{1}{\sigma^2}G_k'(Y-\sum_{j\neq k}G_j\theta_j),
\end{equation}
\begin{equation}\label{Sigmak}
    \hat{\Sigma}_k=(\lambda_k^{-1}K^{-1}+\frac{1}{\sigma^2}G_k'G_k)^{-1}.
\end{equation}
\end{subequations}

The above equations already allow to implement Gibbs sampling 
where the following samples are generated  at any iteration $t$:
\begin{subequations}
\begin{equation}\label{gibbsLambdak}
\begin{split}
   &\lambda_k^{(t)} \mid (Y, \sigma^{2(t-1)},\{\lambda_j^{(t)}\}_{j=1}^{j=k-1}, \{\lambda_j^{(t-1)}\}_{j=k+1}^{j=m},\\ &
   \{\theta_j^{(t-1)}\} ),  \quad \quad \text{for}~k=1,\cdots,m,
\end{split}
\end{equation}
\begin{equation}\label{gibbsSigma2}
    \sigma^{2(t)} \mid (Y,\{\lambda_j^{(t)}\},\{\theta_j^{(t-1)}\}),
\end{equation}
\begin{equation}\label{gibbsThetak_d}
\begin{split}
    \theta_k^{(t)} \mid (Y,\sigma^{2(t)},\{\lambda_j^{(t)}\}, \{\theta_j^{(t)}\}_{j=1}^{k-1}, \{\theta_j^{(t-1)}\}_{j=k+1}^{m}),&\\
    \text{for}~k=1,\cdots,m.&
\end{split}
\end{equation}
\end{subequations}

However, as anticipated, such classical approach
may have difficulties to handle collinearity. 
A simple variation consists of a random sweep Gibbs sampling scheme.
It updates the blocks $\{\theta_k^{(t)}\}$ randomly, trying to find a strategy to increase the mixing.
But it will be shown that this change is not significant enough using numerical examples in Section~\ref{secExp}.
This motivates the development described in the next sections.

\subsection{Overlapping blocks}\label{subsecOB}
\begin{figure}[!ht]
\centering
\includegraphics[scale=0.6]{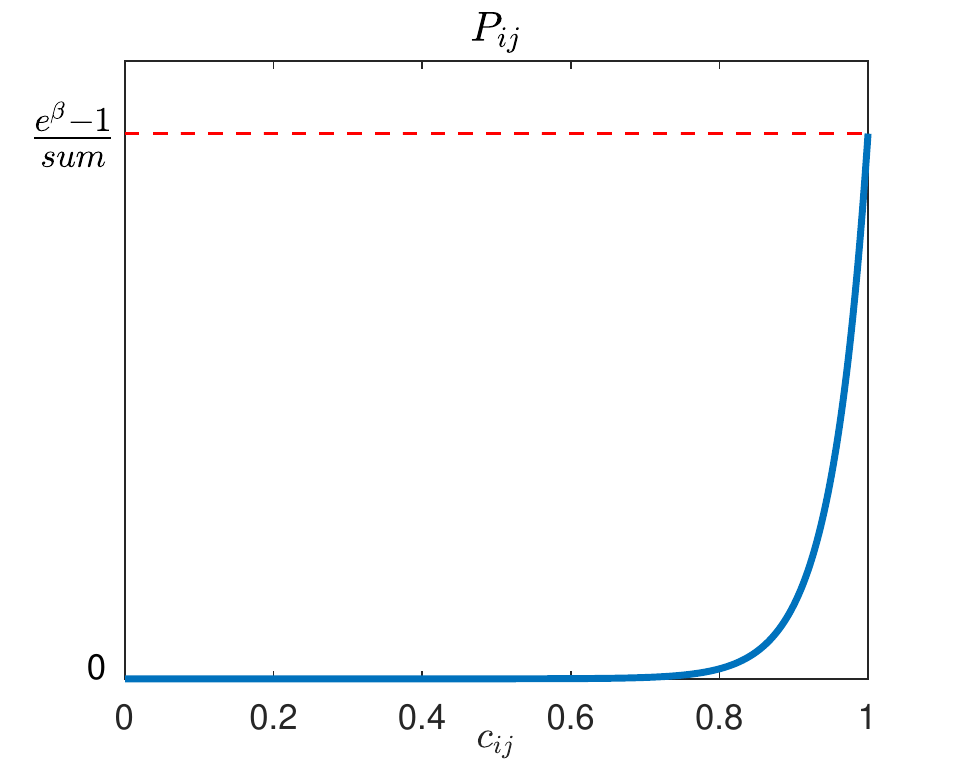}
\caption{Probability $P_{ij}$ related to the frequency with which the scheme updates two impulse responses as a function of the index $c_{ij}$ related to their
posterior correlation.}
\label{cp}
\end{figure}

In our system identification setting, collinearity is related to the
relationship between two inputs $u_i$ and $u_j$ which can make some posterior regions
difficult to explore. A collinearity measure is now needed to drive
the MCMC scheme more towards such parts of the parameter space.
In particular, beyond sampling the single $\theta_i$
through the full conditional distributions \eqref{theta_k},
it is crucial also
to update larger blocks. Such blocks should contain e.g. couples of impulse responses
with high posterior correlation.
Let the inputs $u_i$ be stochastic processes. Then, a first important index is the absolute value of the correlation coefficient, i.e.
\begin{equation}\label{cij}
    c_{ij}:=\left| \frac{\text{Cov}(u_i,u_j)}{\sqrt{\text{Var}(u_i)\text{Var}(u_j)}}\right|,
\end{equation}
where $\text{Cov} (\cdot,\cdot)$ and $\text{Var}(\cdot)$ denote the covariance and variance, respectively.
If the inputs are not stochastic or their statistics are unknown,
such index can be still computed just replacing the variance and covariance in \eqref{cij}
with their estimates, i.e. the sample variance and covariance.\\
\indent It is now crucial to define a suitable function which maps
each $c_{ij}$  into a probability related to the frequency with which the scheme updates the vector containing both $\theta_i$ and $\theta_j$ at once.
First, it is useful to define $\theta_{ij}:= [\theta_i', \theta_j']'$ and let $P_{ij}$ be the probability connected
with the frequency of updating $\theta_{ij}$ inside an MCMC iteration.
Then, we  use an exponential rule
that emphasises  correlation coefficients close to one, i.e.
\begin{equation}\label{Pij}
    P_{ij}=\left\{\begin{array}{cc}
               0,& i=j;  \\
               \frac{e^{\beta c_{ij}}-1}{sum}, &i\neq j;
             \end{array}\right.
\end{equation}
where $sum=\sum_{i<j}(e^{\beta c_{ij}}-1)$ while $\beta$ is a tuning rate.
As an illustrative example, the profile of $P_{ij}$ as a function of $c_{ij}$ with $\beta=20$
is visible in Fig.~\ref{cp}.

Letting $G_{ij}=[G_i,G_j]$, the conditional distribution of $\theta_{ij}$ is
\begin{subequations}\label{BlockUpdate}
\begin{equation}\label{thetaij}
    \theta_{ij}\mid (Y, \sigma^{2},\{\theta_k\}_{k\neq i,j}, \{\lambda_k\}) \sim \mathcal{N}(\hat{\mu}_{ij}, \hat{\Sigma}_{ij})
\end{equation}
where
\begin{equation}\label{Eij}
    \hat{\Sigma}_{ij}=(\begin{bmatrix}\lambda_i^{-1}K^{-1}&0\\
    0 & \lambda_j^{-1}K^{-1}\end{bmatrix}
    +\frac{1}{\sigma^2}G_{ij}'G_{ij})^{-1},
\end{equation}
\begin{equation}\label{muij}
      \hat{\mu}_{ij}= \hat{\Sigma}_{ij}\frac{1}{\sigma^2}G_{ij}'(Y-\sum_{k\neq i,j} G_k\theta_k).
\end{equation}
\end{subequations}
The above equations may thus complement the Gibbs sampling scheme previously described.
They allow to update also additional overlapping blocks consisting
of larger pools of impulse responses with a frequency connected to $P_{ij}$.

\subsection{The role of the stable spline kernel scale factors and the use of a common one}
The simulation strategy described in the previous section uses a model
where the covariance of each impulse response depends on a
different stochastic scale factor $\lambda_k$.
As illustrated via numerical experiments in Section~\ref{secExp},
this modeling choice may have a harmful effect in presence of strong collinearity.
In fact, it can make the generated chain nearly reducible \cite[Chapter 3]{Gilks}, 
i.e. unable to visit the highly correlated parts of the posterior. 
The problem can be overcome by reducing the model complexity,
i.e. assigning a common scale factor to all the impulse responses covariances.
This can appear a paradox since one could think that a model with more parameters
may permit the chain to move more easily along the posterior's support.
Instead, the introduction of many scale factors can trap the algorithm
in regions from which it is virtually impossible to escape.\\
Using only one common scale factor for all the $\theta_k$ covariances, the distribution \eqref{postDisLambdak} becomes
\begin{equation}\label{postDisLambda}
    \lambda \mid (Y, \sigma^2, \{\theta_k\}) \sim {\mathcal{I}_g}(\frac{np+1}{2},~ \frac{1}{2}\sum \theta_k'K^{-1}\theta_k)
\end{equation}
and the update of $\lambda$, previously defined by \eqref{gibbsLambdak}, is given by
\begin{equation}\label{gibbsLambda}
    \lambda^{(t)} \mid (Y, \sigma^{2(t-1)}, \{\theta_k^{(t-1)}\} ).
\end{equation}
In turn, at step $t$ of the MCMC algorithm, in place of
differently from \eqref{gibbsThetak_d}, the algorithm will sample $\theta_k$ using 
\begin{equation}\label{gibbsThetak}
   \theta_k^{(t)} \mid (Y,\sigma^{2(t)}, \lambda^{(t)}, \{\theta_j^{(t)}\}_{j=1}^{k-1}, \{\theta_j^{(t-1)}\}_{j=k+1}^{m}).
\end{equation}
Finally, when the block $\theta_{ij}$ is selected, the update rule becomes 
\begin{equation}\label{gibbsThetaij}
     \theta_{ij}^{(t)} \mid (Y, \sigma^{2(t)},\lambda^{(t)}, \{\theta_k^{(t)}\}_{k\neq i,j}),
\end{equation}
i.e. \eqref{BlockUpdate} is now used with $\lambda_i=\lambda_j=\lambda^{(t)}$.

\subsection{Random sweep sampling schemes: the RSGSOB algorithm}

By working upon all the developments previously described,
in this section we finalize our MCMC strategy for linear system identification under collinear inputs.
Besides updating the scale factor and noise variance using \eqref{gibbsLambda} and
\eqref{gibbsSigma2}, we have also to define the exact rule to update the blocks and overlapping blocks
during any iteration. For this purpose, we will extend the so called random sweep
Gibbs sampling scheme including overlapping blocks. The main idea is that at any iteration
only one of the blocks  $\{\theta_i,\theta_{ij}\}$ is updated, chosen according to a discrete probability density
function of the scalars $P_{ij}$
introduced in Section~\ref{subsecOB}. The symmetries $c_{ij}=c_{ji}$, $P_{ij}=P_{ji}$, and the fact that
$\theta_{ij}$ and $\theta_{ji}$ contain the same impulse responses coefficients, imply that
only the blocks $\theta_{ij}$ with $i<j$ need to be considered in what follows.\\
Let $\mathcal{M}_1:=\{\theta_i\}$ and $\mathcal{M}_2:=\{\theta_{ij}\}_{i<j}$ and define
$$
    \mathcal{M}:= \mathcal{M}_1 \bigcup \mathcal{M}_2.
$$
Now, we specify for any block in $\mathcal{M}$ the probability of being selected by the random sweep sampling scheme.
It is natural to assign the same probability to the blocks in $\mathcal{M}_1$.
For what concerns the overlapping ones, we need to take into account the collinearity indexes discussed in Section~\ref{subsecOB}.
Overall, the discrete probability density $P_\mathcal{M}$ over $\mathcal{M}$ is then defined by
\begin{equation}\label{PM}
    P_\mathcal{M}(b)=\left\{\begin{array}{cc}
            \frac{1}{m+n_{OB}}, &  \text{if} \ b=\theta_i\in \mathcal{M}_1,\\
            \frac{n_{OB}}{m+n_{OB}}P_{ij}, &  \text{if} \ b=\theta_{ij}\in \mathcal{M}_2,
    \end{array}\right.
\end{equation}
so that $\sum_{\mathcal{M}}P_\mathcal{M}(b)=1$ for any
$n_{OB}\in \mathbb{Z}_+$ which represents a tuning parameter.
It regulates how frequently we want to consider the collinearity problem during our sampling.\\

Finally, our random sweep
Gibbs samplings with overlapping blocks
is summarized in Algorithm~\ref{algRSGSOB}
where  $n_{MC}$ indicates the number of MCMC steps while $n_B$ is the length of the burn-in \cite[Chapter 7]{Gilks}.
The procedure is called RSGSOB in what follows.

\begin{figure}
\begin{algorithm}[H]
\caption{Random sweep Gibbs samplings with overlapping blocks using one
    common scale factor ({\bf RSGSOB})}\label{algRSGSOB}
\begin{algorithmic}[1]
\REQUIRE Measurements $G$, $Y$; initial values $\lambda^{(0)}$, $\sigma^{2(0)}$, $\{\theta_k^{(0)}\}$; $\alpha$, $\beta$, $n_{OB}$, $n_B$.
      \ENSURE Estimate $\hat\theta$.
      \STATE Calculate $\{P_{ij}\}$ from input data for $i\leq j$ in \eqref{Pij};
      \FOR {$t=1:n_{MC}$}
      \STATE Sample \eqref{gibbsLambda}, \eqref{gibbsSigma2} in sequence;
        \FOR {$s=1:m+ n_{OB}$}
        \STATE Choose a block $b$ from the distribution $\{P_\mathcal{M}\}$;
            \IF{$b=\theta_i\in \mathcal{M}_1$}
                \STATE Sample \eqref{theta_k} given the latest data of $\theta$ and update the values of $\theta_i^{(t)}$;
            \ELSE[$b=\theta_{ij}\in \mathcal{M}_2$]
                \STATE  Sample \eqref{gibbsThetaij} given the latest data of $\theta^{(t)}$ and update the values of $\theta_i^{(t)}$ and $\theta_j^{(t)}$.
            \ENDIF
        \ENDFOR
      \ENDFOR
     \STATE Calculate $\hat\theta$ as the mean of the impulse responses samples from $t=n_B+1$ to $t=n_{MC}$.
\end{algorithmic}
\end{algorithm}
\end{figure}

%
%

\section{Convergence analysis}\label{secConvg}

\subsection{The main convergence theorem}

In this section we show that RSGSOB generates a Markov chain convergent to the desired posterior
of impulse responses and hyperparameters. In addition,
the speed of convergence is characterized.
Finally, a simple example is reported to illustrate the practical
implications of our theoretical findings.\\
We consider the convergence rate in the $L^2$ sense
in the setting of spectral analysis, as initially proposed by \cite{Goodman89}.
This includes the study of how quickly the expectations of arbitrary square
$h$-integrable functions approach their stationary values.
Since $\theta$ is the parameter really of interest to estimate,
we focus on the convergence rate of the main part in Algorithm~\ref{algRSGSOB}, i.e., from line $4$ to line $11$.
It is then natural to introduce the following definition.

\begin{defn}[Convergence rate]\label{def1}
Let $h^{(t)}(\theta \mid \theta^{(0)})$ denote the  probability density of a Markov chain at time $t$ given the starting point $\theta^{(0)}$.  Assume that the Markov chain converges to the probability density function $h$
and denote by $\mathbb{E}_{h(\theta)}[f(\theta)]$ the expectation of a real-valued function $f$ under $h$ with respect to $\theta$, i.e.,
$$
    \mathbb{E}_{h(\theta)}[f(\theta)]= \int f(\theta) h(\theta){\rm d}\theta.
$$
Let
$$
    P^t(f):=\int f(\theta)h^{(t)}(\theta \mid \theta^{(0)}) {\rm d}\theta
$$
describe  the expectation of $f(\theta)$ with respect to the density of the Markov chain at time $t$ given $\theta^{(0)}$.
Then, the convergence rate of $\theta$ given $\theta^{(0)}$ is the minimum scalar $rate$ such that for all
the square $h$-integrable functions $f$,
and for all $r>rate$, one has
\begin{equation}\label{CRdef}
  \lim_{t\rightarrow \infty}~\mathbb{E}_{h(\theta^{(0)})} \left[ \{ P^t(f)-\mathbb{E}_{h(\theta)}[f(\theta)]\}^2 \right]r^{-t} =0.
\end{equation}
Hence, $rate$ quantifies how fast the density of the Markov chain approaches the true distribution in $L^2$ as
$t$ grows to infinity.
\end{defn}

Note that. according to the above definition, the convergence rate is independent of the initial chain value.
Additional details on the nature of this notion of convergence can be found in \cite{RS97}, where the rates are derived for different Gibbs sampling schemes involving Gaussian distributions.\\
Our convergence result, whose proof can be found in Appendix, 
is reported below. To formulate it, we use $\rho(\cdot)$ to denote the spectral radius, i.e., the maximum eigenvalue modulus of a matrix,
while the relevant probability distributions contained in its statement are calculated in Appendix. 
\medskip
\begin{thm}[Convergence of RSGSOB]\label{thmcon}
The RSGSOB scheme described in Algorithm 1 generates a Markov chain convergent
to the correct posterior of impulse responses and hyperparameters in accordance with the Bayesian
model for linear system identification described in Section \ref{secProb}.
In addition, for fixed hyperparameters $\lambda$ and $\sigma^2$, 
the convergence rate in Definition~\ref{def1} 
is
\begin{equation}\label{rate}
   rate({\rm RSGSOB})=\rho(C^{m+n_{OB}})=\rho(C)^{m+n_{OB}},
\end{equation}
where
\begin{equation}\label{Csum}
\begin{split}
    C :=&\sum_{\mathcal{M}_1}P_{\mathcal{M}}(\theta_i)C_i + \sum_{\mathcal{M}_2}P_{\mathcal{M}}(\theta_{ij})C_{ij}\\
     =& \frac{1}{m+n_{OB}} (\sum_{\mathcal{M}_1}C_i + n_{OB} \sum_{\mathcal{M}_2}P_{ij}C_{ij}),
\end{split}
\end{equation}
the matrices $C_i$ and $C_{ij}$ are reported, respectively, in \eqref{Ci} and \eqref{Cij} with $\lambda^{(t)}=\lambda$ and $\sigma^{2(t)}=\sigma^2$.
\end{thm}

\subsection{A simple example}

While more sophisticated system identification problems are discussed in the next section,
a simple example is now introduced to show that the use of overlapping blocks,
the key feature of RSGSOB, can fasten the convergence in comparison with
a classical random sweep Gibbs sampling (RSGS) scheme.
This latter procedure, also described in the next section in TABLE~\ref{table_ex1algs},
samples the blocks $\theta_i$ with the same probability ${1}/{m}$ for $m+n_{OB}$ times
during any iteration.
Its convergence rate is obtained below using Lemma~\ref{thmRob3} and Lemma~\ref{lem2}
contained in Appendix.
\medskip
\begin{prop}[Convergence rate of RSGS]\label{thmcon2}
With hyperparameters $\lambda$ and $\sigma^2$ fixed, the convergence rate of RSGS is
\begin{equation}\label{rateRSGS}
   rate({\rm RSGS})= \rho(\frac{1}{m}\sum_{i=1}^m C_i )^{m+n_{OB}},
\end{equation}
where the matrix $C_i$ is defined in \eqref{Ci}.
\end{prop}
\medskip
Now, consider a toy and extreme example where all the correlation coefficients
among the inputs are equal to one. In particular, we set
$\lambda=1$ while the noise variance is $\sigma^2=1$ and consider
a system with $10$ inputs, all equal to a Dirac delta function.
It comes that the regression matrix $G$ in \eqref{MM2} has dimension $10 \times 100$
and any $G_i$ is the identity matrix. Setting $n_{OB}=3$, $\alpha=0.9$, and $\beta=100$,
from  \eqref{rate} and \eqref{rateRSGS} one obtains
\begin{align}
    \nonumber
  &rate({\rm RSGSOB}) =  0.5861,\\ \nonumber
  &rate({\rm RSGS}) =0.8045.
\end{align}
This outlines how the use of RSGSOB can greatly enhance the convergence rate of the
MCMC procedure.

\section{Simulation examples}\label{secExp}
\subsection{Example 1: an extreme case}
We first illustrate one extreme case of collinearity where the system is fed with two identical inputs. 
This simple example will point out the importance of introducing overlapping blocks
in the sampling scheme and of adopting only a common scale factor $\lambda$
for all the impulse responses.

\begin{figure*}[!ht]
\centering
\subfigure[GS]{\includegraphics[scale=0.56]{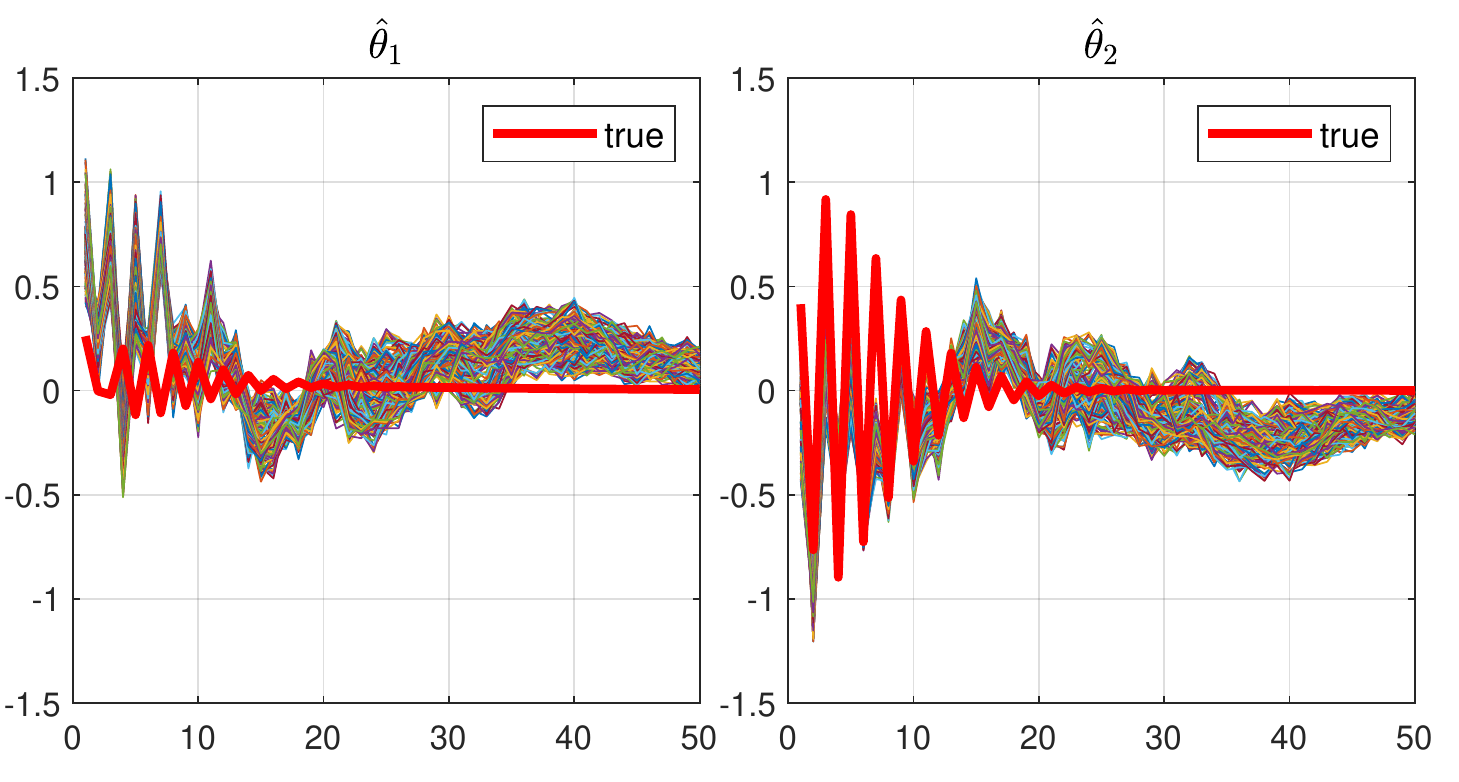}}\hspace{0.8cm}\label{ex1_ir_GS}
\subfigure[GSd]{\includegraphics[scale=0.56]{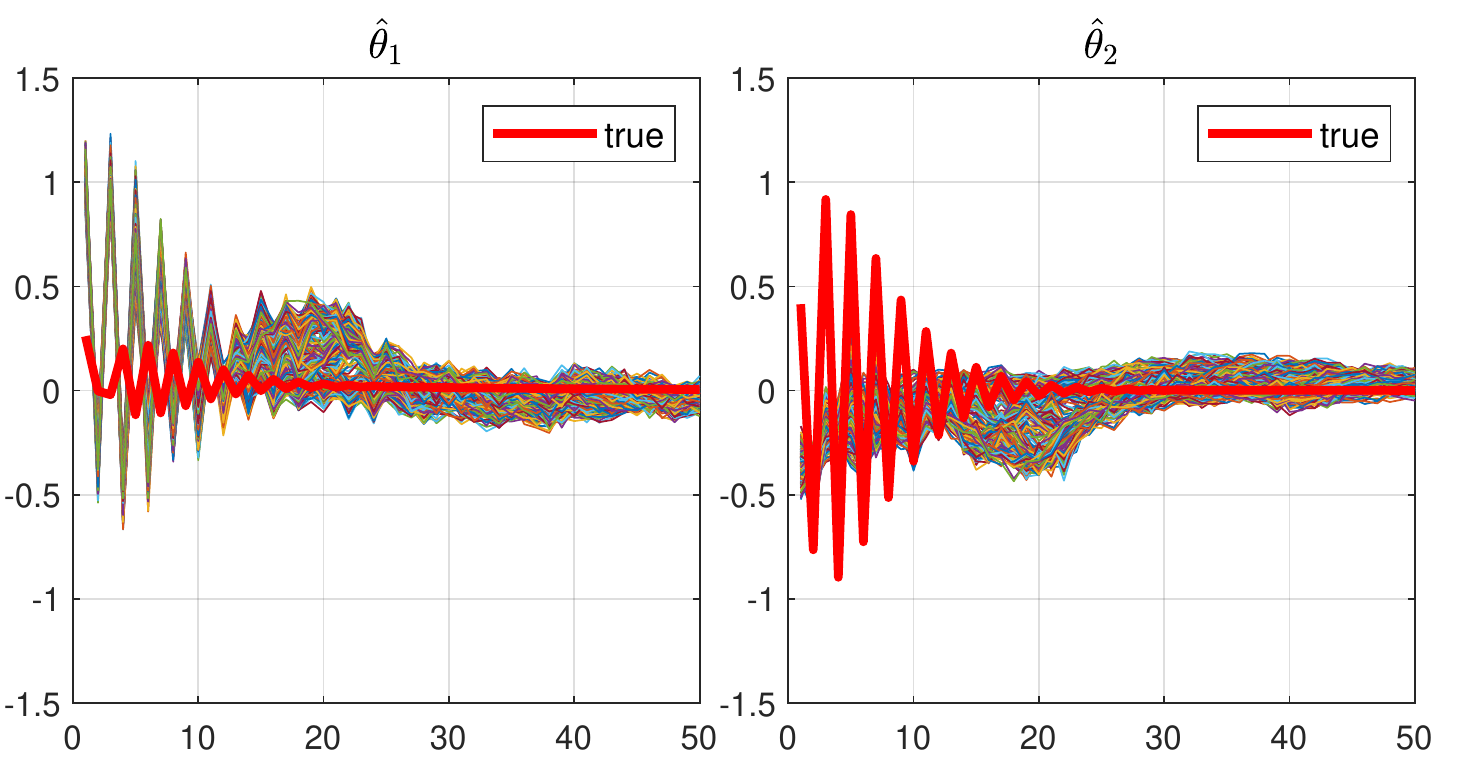}}\label{ex1_ir_GSd}

\subfigure[RSGS]{\includegraphics[scale=0.56]{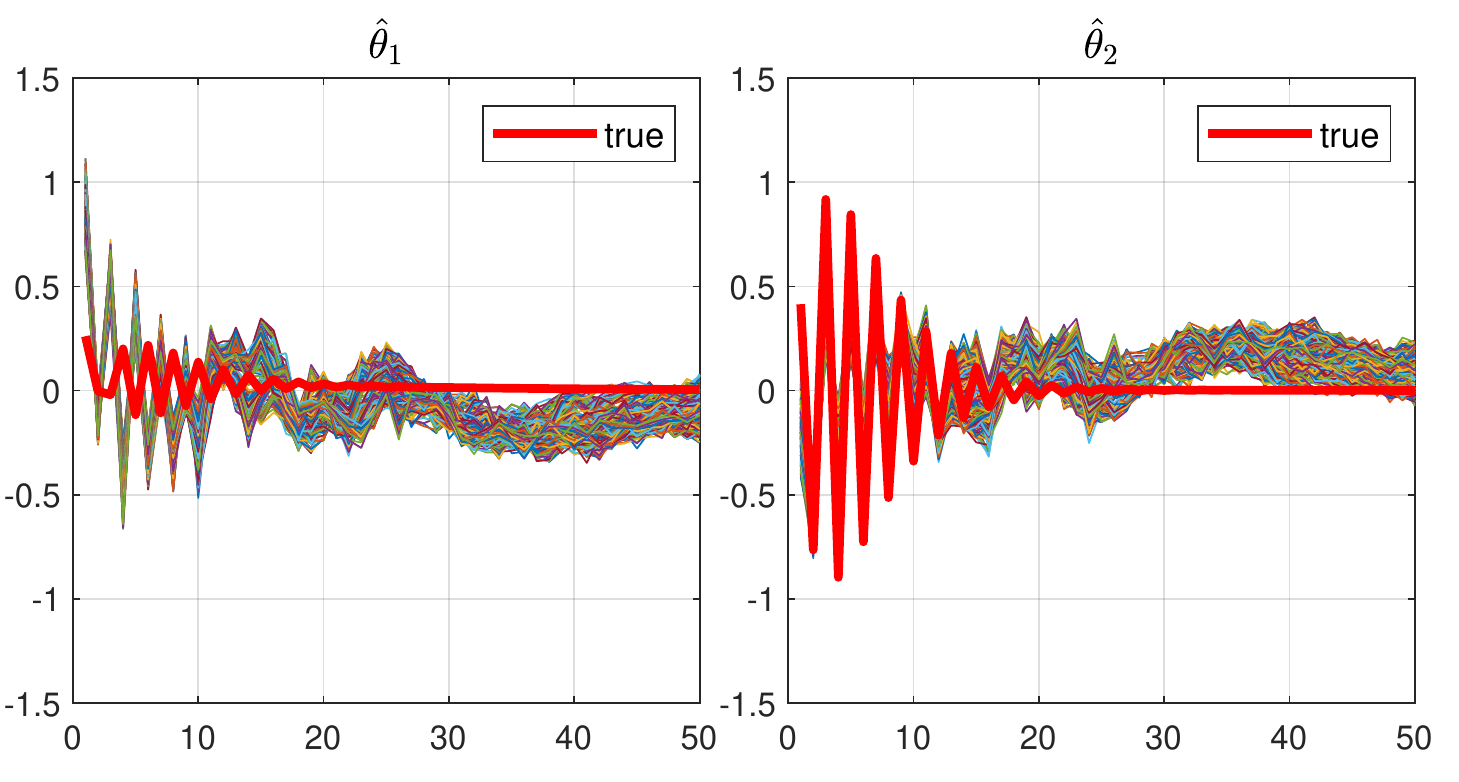}}\label{ex1_ir_RSGS}\hspace{0.8cm}
\subfigure[RSGSd]{\includegraphics[scale=0.56]{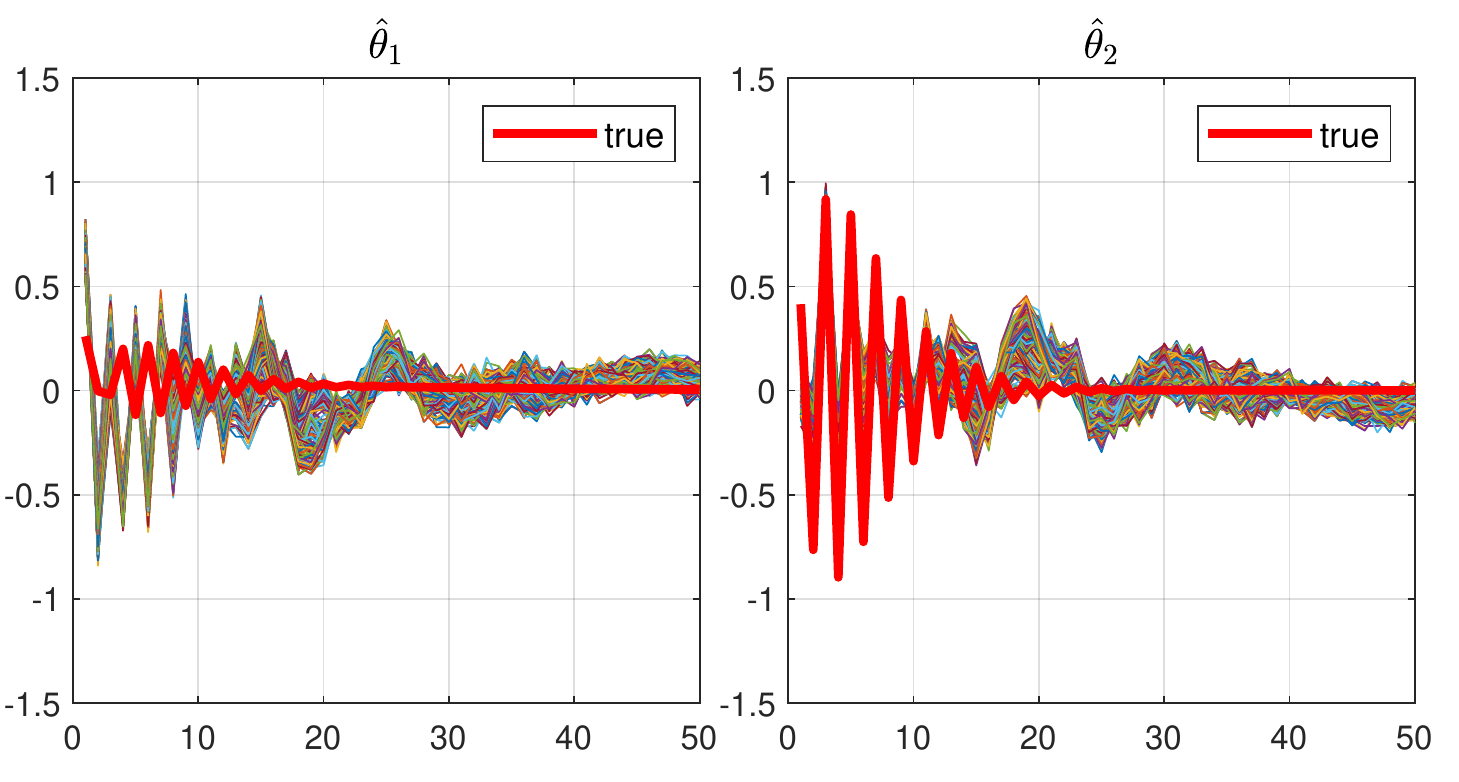}}\label{ex1_ir_RSGSd}

\subfigure[\bf{RSGSOB: Algorithm 1}]{\includegraphics[scale=0.56]{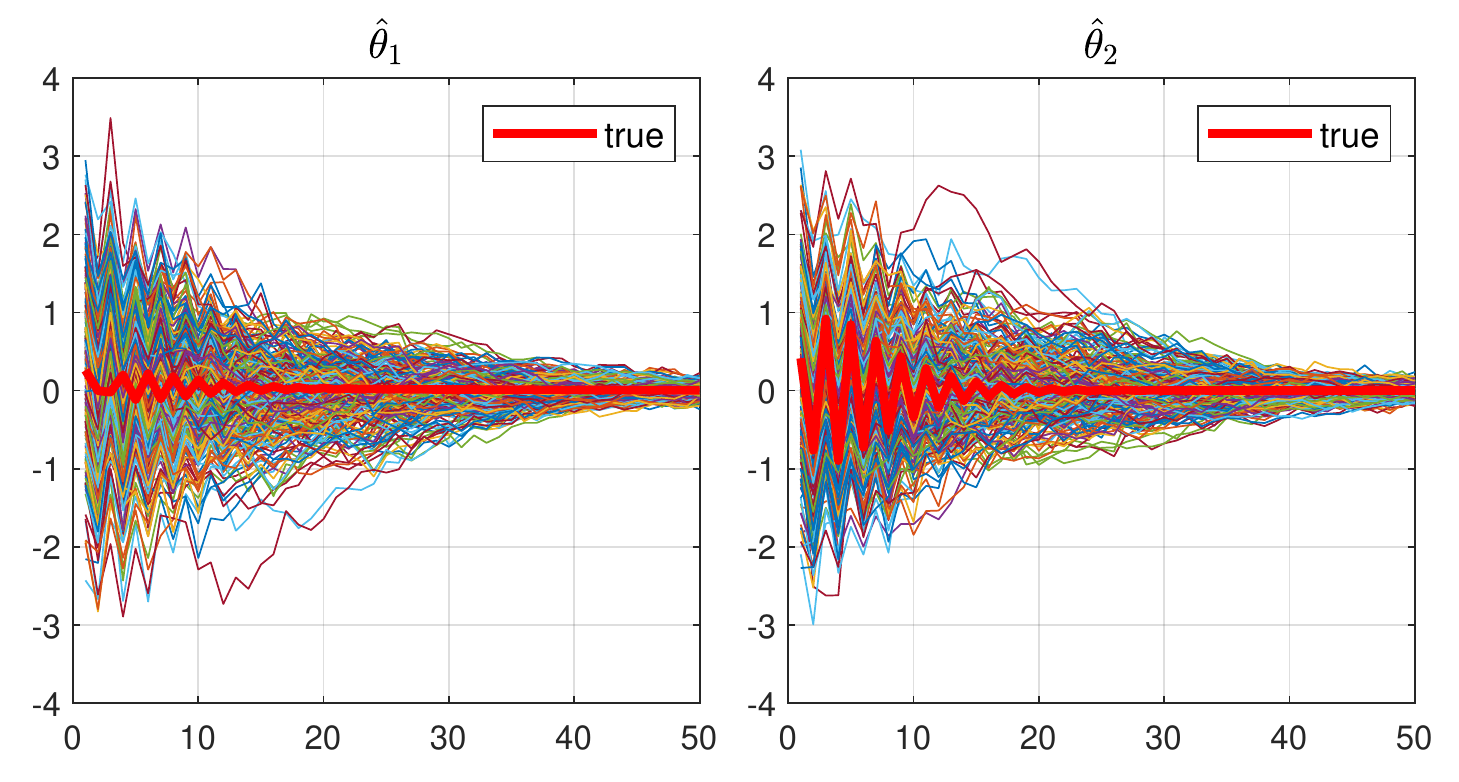}}\label{ex1_ir_RSGSOB}\hspace{0.8cm}
\subfigure[RSGSOBd]{\includegraphics[scale=0.56]{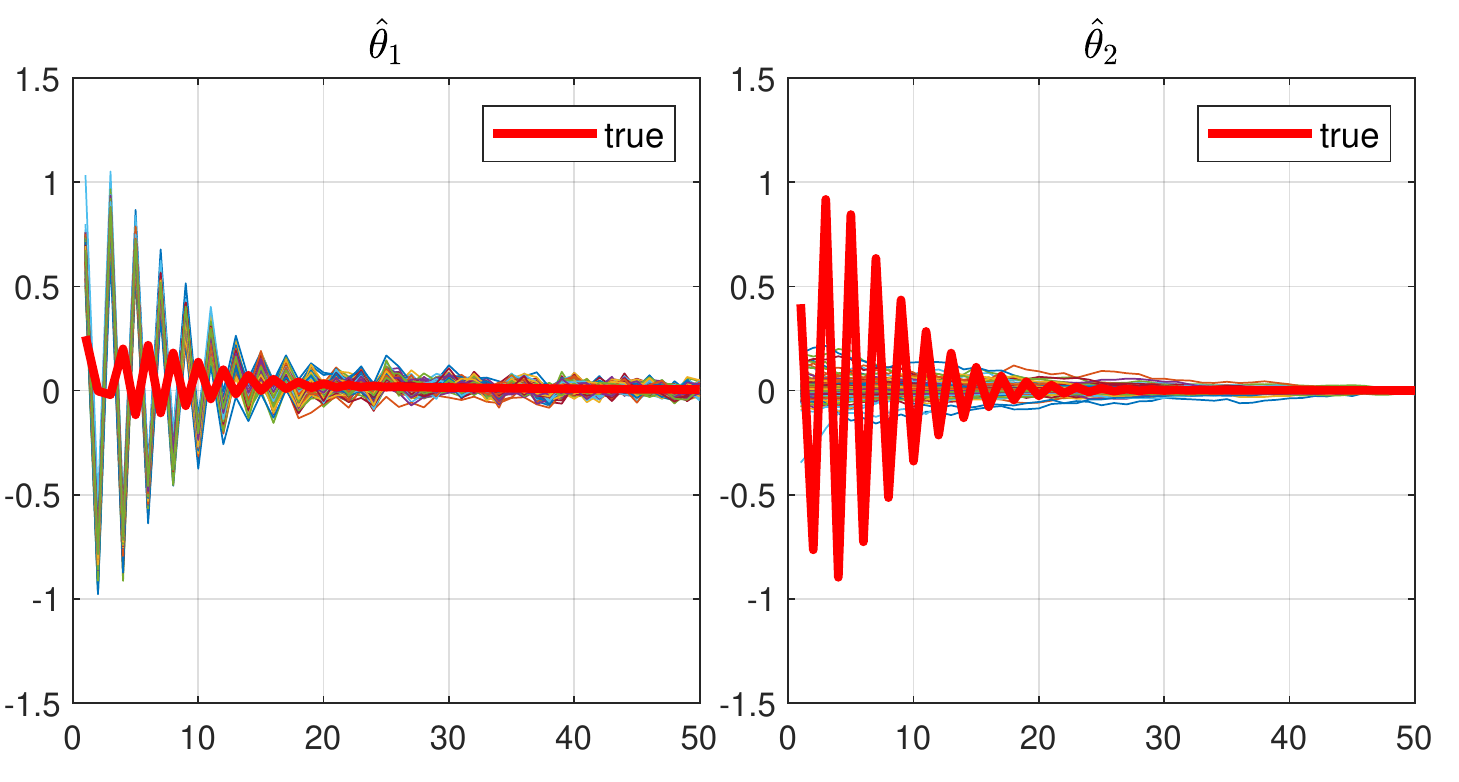}}\label{ex1_ir_RSGSOBd}
\caption{\emph{Example 1}: posterior of the two impulse responses reconstructed in sampled form after 500 iterations by the 6 algorithms
described in Table~\ref{table_ex1algs}.}\label{fig_ex1}
\end{figure*}

Let $u_1(t)=u_2(t)$, with the common input defined by realizations of
white Gaussian noise with $n=500$. Let also $F_1(z), F_2(z)$
be two randomly generated transfer functions with a common denominator of degree 5,
displayed in the panels of Fig. \ref{fig_ex1}.
The measurement noises $e_i$ are independent Gaussians of variance equal to 
the sample variance of $\sum_{k=1}^2 F_k u_k$ divided by 5. 
In the Fisherian framework adopted by classical system identification
impulse responses estimation is a non-identifiable problem.
But we can use the model \eqref{MM2} with the dimension
of $\theta_1$ and $\theta_2$ set to $p=50$ within the Bayesian framework.
The goal is to find the impulse responses posterior under the stable spline prior \eqref{SS} where here, and in
Example 2, the variance decay rate is $\alpha=0.9$.

In this example, one has the number of inputs $m=2$ and
the collinearity index  $c_{12}=1$, so that
$P_{12}=1$.
We also set $n_{OB}=2$ in \eqref{PM} obtaining
 $P_\mathcal{M}(\theta_1)=P_\mathcal{M}(\theta_2)=1/4$, and $P_\mathcal{M}(\theta_{12})=1/2$. 
We run $n_{MC}=500$ Monte-Carlo iterations in MATLAB to compare 6 algorithms described in Table~\ref{table_ex1algs}, where our proposed Algorithm~\ref{algRSGSOB} is  abbreviated to RSGSOB, and highlighted bold, while RSGSOBd is the version which adopts
different scale factors, one for any stable spline kernel modeling $\theta_k$.
The other algorithms include Gibbs sampling with one or multiple scale factors, respectively called GS and GSd,
Random sweep Gibbs sampling with one ore multiple scale factors, respectively called RSGS and RSGSd.

\begin{table*}[!t]
\centering
\caption{6 algorithms to compare in Example 1}\label{table_ex1algs}
\begin{tabular}{ccc}
  \toprule
  \bf{Abbr.} & \bf{Algorithm} &\bf{Update in one iteration} \\ \toprule
  GS & Gibbs sampling&  \eqref{gibbsLambda}\eqref{gibbsSigma2}\eqref{gibbsThetak} in sequence \\ \hline
  GSd & GS using different scale factors & \eqref{gibbsLambdak}\eqref{gibbsSigma2}\eqref{gibbsThetak_d}  in sequence \\ \hline
  RSGS & Random sweep Gibbs sampling &  \begin{tabular}{c} first \eqref{gibbsLambda}\eqref{gibbsSigma2}, then repeat $m+n_{OB}$ times:\\ randomly choose and update a block in $\{\theta_i\}$\\ with the same probability $1/m$ \end{tabular} \\ \hline
  RSGSd & RSGS using different scale factors & \begin{tabular}{c}the same as RSGS,\\ except for updating \eqref{gibbsLambdak} instead of \eqref{gibbsLambda}\end{tabular} \\ \hline
  \bf{RSGSOB} & \begin{tabular}{c}RSGS with overlapping blocks\\
  (\bf{Algorithm~\ref{algRSGSOB}})\end{tabular} & line 3 to line 11 in Algorithm~\ref{algRSGSOB} \\ \hline
  RSGSOBd & RSGSOB using different scale factors & \begin{tabular}{c}the same as RSGSOB,\\ except for updating \eqref{gibbsLambdak} instead of \eqref{gibbsLambda}\end{tabular} \\
  \bottomrule
\end{tabular}
\end{table*}

The posterior of the two impulse responses is reconstructed 
using samples from MCMC considering the first $50\%$ as burn-in. 
Results coming from the six procedures are displayed in Fig.~\ref{fig_ex1}, where true impulse responses are the red thick lines. For clarity, the scale of the vertical axis of Fig.~\ref{fig_ex1}(e) is different from other sub-figures in Fig.~\ref{fig_ex1}.

Let us start commenting the performance of the algorithms using only one common scale factor.
Results show that only RSGSOB 
returns an informative posterior, able to highlight the non-identifiability issues, see Figs.~\ref{fig_ex1}(e).
Results from classical Gibbs sampling (GS) and Random sweep Gibbs sampling (RSGS) are in Figs.~\ref{fig_ex1}(a) and~\ref{fig_ex1}(c).
These two algorithms 
have similar performance, not significantly affected by choosing deterministic or random sampling scheme
(this also holds in Example 2, so that we will hereby only use RSGS in Table~\ref{table_ex1algs} as a benchmark).
As a matter of fact, both of them 
have not reached the level of information on the posterior shape returned by RSGSOB since
they converge much slower to the target density. 
The use of GS or RSGS, e.g. for control or prediction purposes, can thus lead to erroneous results,
possibly affected by huge errors if future inputs are not very similar.
In addition, differently from RSGSOB, they can largely underestimate the uncertainty around the predictions.

For what regards the algorithms that use multiple scale factors, their performance is really poor.
Figs.~\ref{fig_ex1}(b) and \ref{fig_ex1}(f) show that the samples of the second impulse response $\theta_2$ are close to zero. The reason is that chains with multiple scale factors may easily become nearly reducible.
In fact, during the first MCMC iterations, the algorithm generates an impulse response $\theta_1$ able to explain the data.
Hence, $\theta_2$ is virtually set to zero, so that also $\lambda_2$ becomes very small.
This scale factor becomes a strong (and wrong) prior for $\theta_2$ during all the subsequent iterations.
This in practice forces the sampler to always use only $\theta_1$ to describe the experimental evidence.
A minor phenomenon to mention is that one can see from Figs.~\ref{fig_ex1}(c) and \ref{fig_ex1}(d) that random sweep Gibbs sampling,
which does not use overlapping blocks, decreases the negative influence of using different scale factors, but remains affected by slow mixing.

\subsection{Example 2: identification of large-scale systems}
Now,  we use 100 inputs to test the algorithms in  large-scale systems identification.
This is a situation where it is crucial to divide the parameter space into many small blocks for computational reasons.
Otherwise, at any MCMC iteration one should multiply and invert large matrices whose dimension
depends on the overall number of impulse response coefficients.\\

The collinear inputs are generated as follows
\begin{equation}\label{col}
 u_j(t)=u_i(t)+r_{ij}(t),
\end{equation}
where $r_{ij}(t)$ is a noise independent of $u_i(t)$.
To increase collinearity between inputs and also the level of ill-conditioning affecting the problem,
$r_{ij}(t)$ in \eqref{col} is a moving average (MA) process,
\begin{equation}\nonumber
    r_{ij}(t)=v_{ij}(t)-0.8v_{ij}(t-1),
\end{equation}
where $v_{ij}(t) \sim \mathcal{N}(0,\omega^2_{ij})$.
From \eqref{cij}, one has
\begin{equation}\nonumber
    c_{ij}=\frac{\left| \eta_i \right|}{\sqrt{\eta_i^2 +\omega^2_{ij}/(1-0.8^2)}},
\end{equation}
when $u_i(t)$ is a zero mean process with variance $\eta^2_i$. 
Hence, one can see that the amount of collinearity can be tuned by $\omega_{ij}$.

The data set size is $n=10^5$. We introduce correlation
among 
the first 10 inputs by letting 
\begin{equation}\label{colt1}
    u_{i+1}(t)=u_{i}(t) + r_{(i+1)i}(t),
\end{equation}
for $i=1,\cdots,9$, where the first input $u_1(t)$ is white and Gaussian of variance 1, $r_{(i+1)i}(t)$ is obtained by setting the $c_{i(i+1)}=0.99$ for $i=1,2,\cdots,9$. The other $90$ inputs, i.e. $\{u_i\}_{i=11}^{100}$ are zero-mean standard i.i.d. Gaussian processes of variance $1$.

\begin{figure}[!ht]
\centering
\includegraphics[scale=0.7]{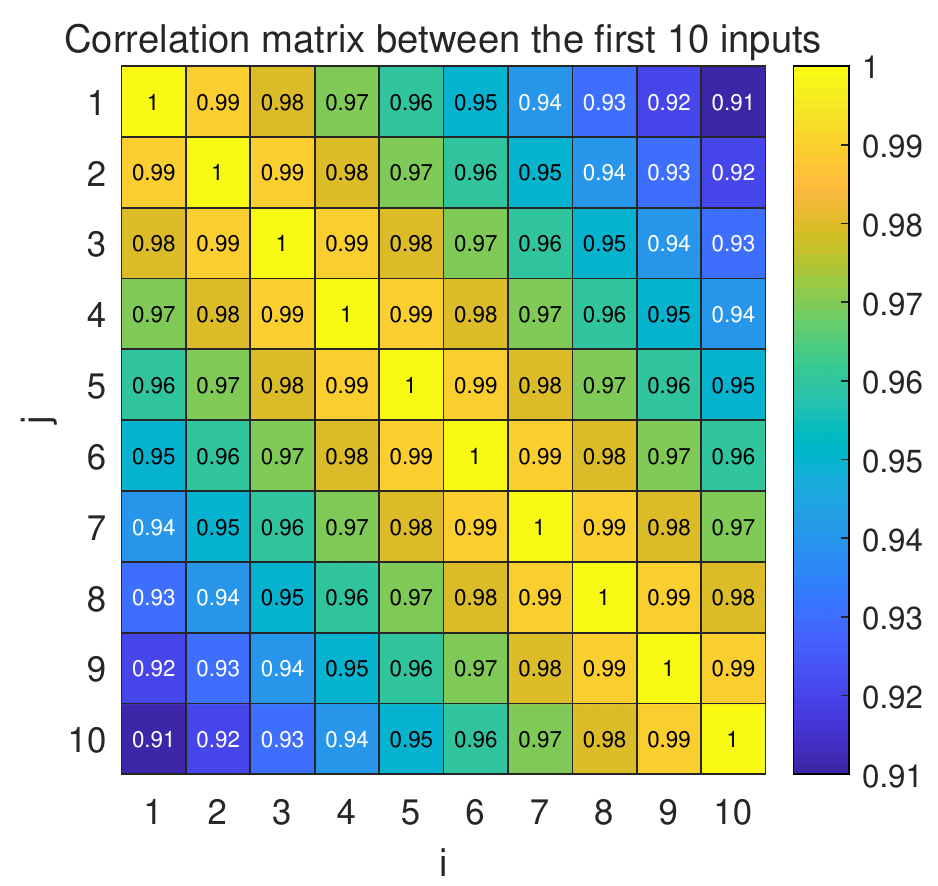}
\caption{\emph{Example 2}: collinearity indexes $\{c_{ij}\}$ of the input couples $\{(u_i,u_j)\}_{i,j=1}^{10}$.}
\label{heatmapCM}
\end{figure}

The different collinearity levels of the first 10 inputs are summarized by the correlation coefficient matrix (calculated from the input realizations) reported in Fig.~\ref{heatmapCM}.
One can see that the input pairs have correlation coefficients ranging from about 0.99 to about 0.91.

\begin{figure*}[htbp]
\centering
\subfigure[Probability level $P_{ij}$ associated to the frequency of choosing
overlapping blocks for $\{\theta_i\}_{i=1}^{10}$ when adopting RSGSOB.]{
\includegraphics[scale=0.56]{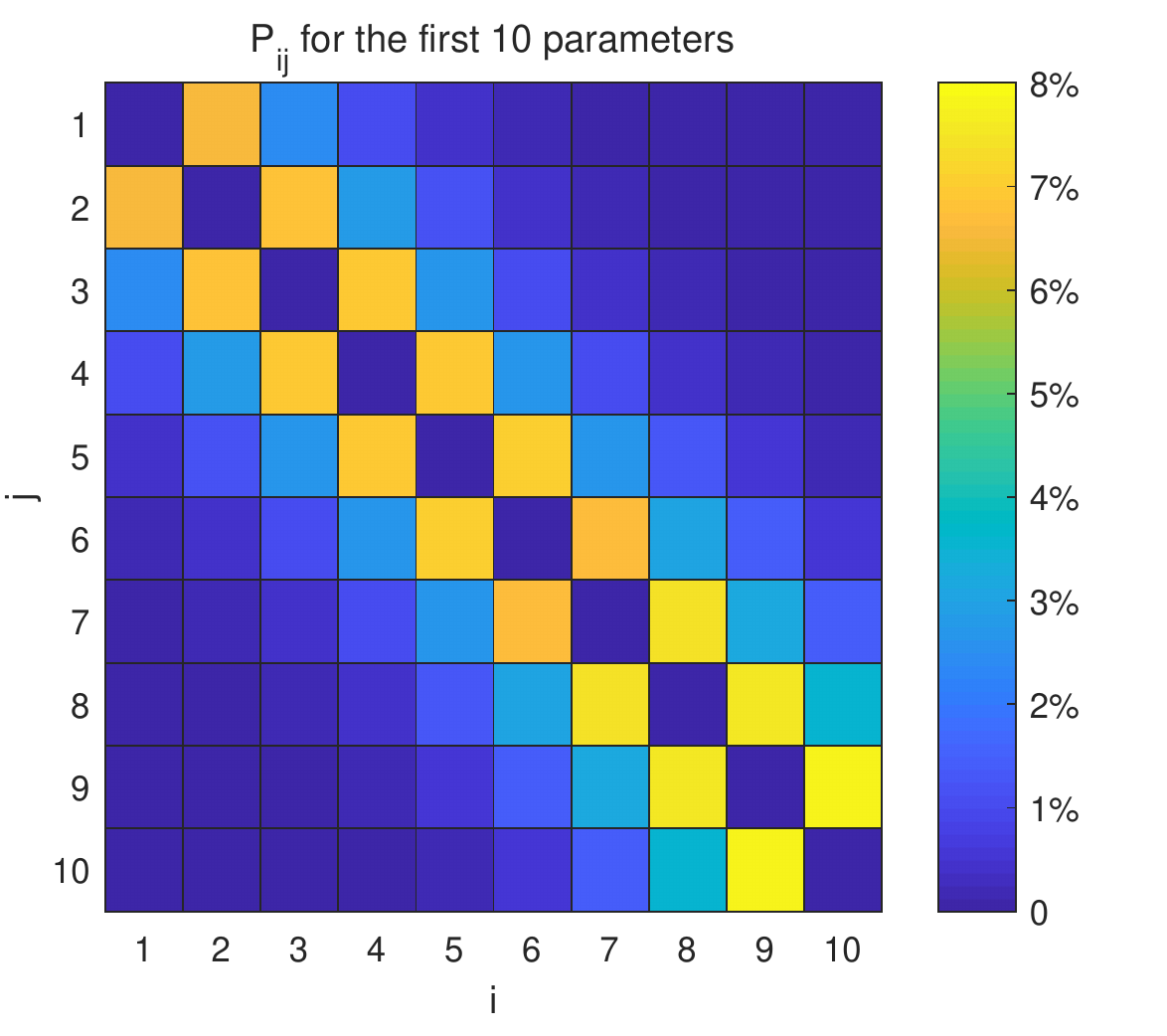}}\label{Pij_ex2}\hspace{0.3cm}
\subfigure[Pie charts of the chosen  blocks. {\bf Left}: frequencies with which the overlapping blocks $b\in \mathcal{M}_2$ have been selected by RSGSOB. {\bf Right}: frequencies with which all the blocks $b\in \mathcal{M}$ have been selected by RSGSOB. ]{\includegraphics[scale=0.56]{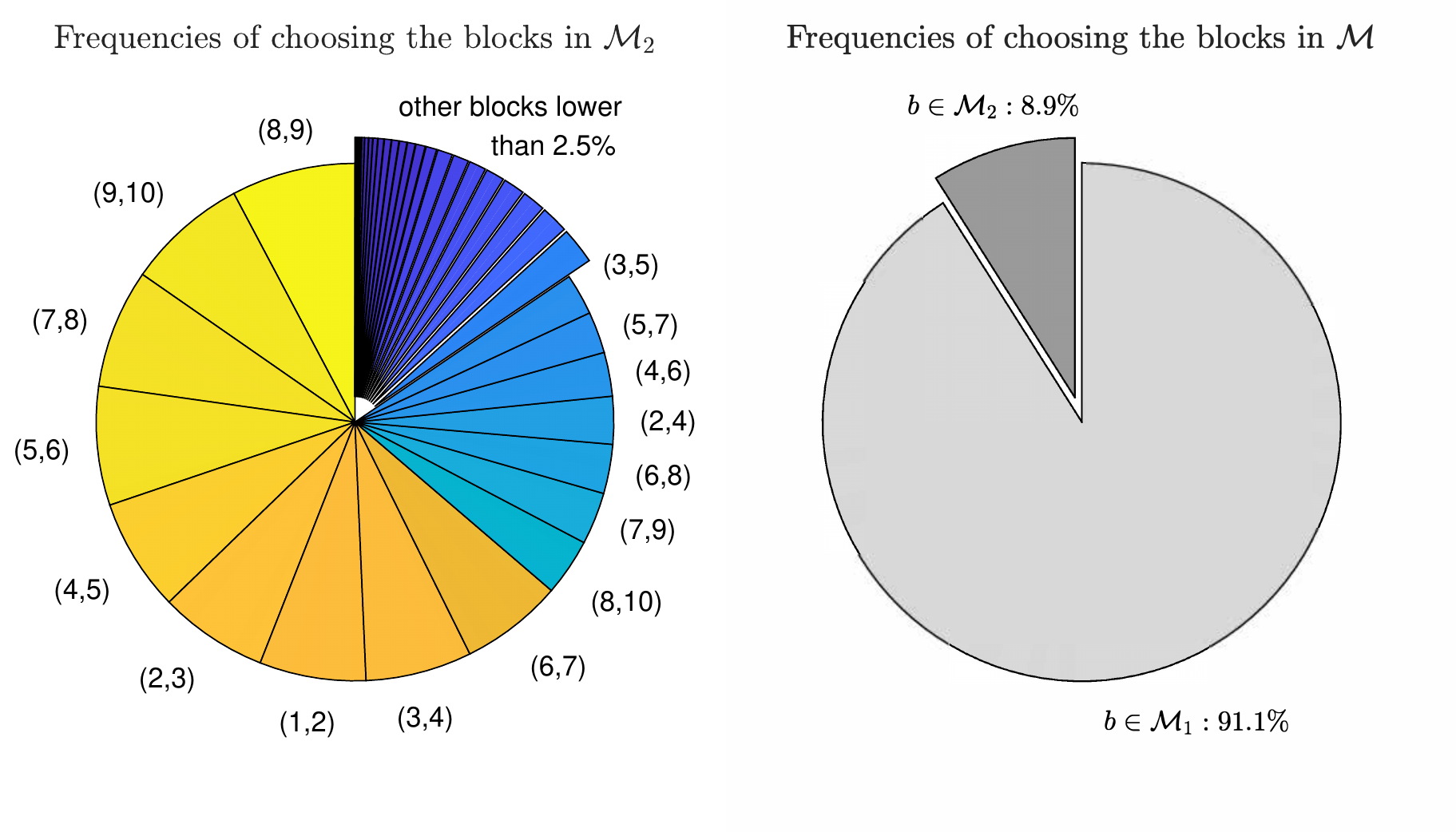}}
\label{Pies_ex2}
\caption{Statistical indexes useful to evaluate how RSGSOB have sampled the different
impulse responses blocks to deal with collinearity in Example 2.}\label{fig_ex2_prob}
\end{figure*}

We set $\beta=100$ and use \eqref{Pij} 
to define the probabilities of selecting the overlapping blocks at any iteration.
The $P_{ij}$ related to the first 10 impulse responses are shown in Fig. \eqref{fig_ex2_prob}
(a) with a sharing colorbar. 
The $P_{ij}$ for  $i,j>10$ are obviously all close to zero: the associated impulse responses couples
do not have significant correlation.
Instead, inside all the pairs $(i,j) (i\neq j)$ considered for constructing overlapping blocks,
the couples $(i,i+1)$ ($i=1,...,9$) of highest collinearity are assigned near $7\%$ of the total amount of probability.
The couple $(1,10)$, which has a correlation coefficient $0.9135$, is given $0.0063\%$.

We generate random transfer functions $F_i(z)~(i=1,\cdots, 100)$ of degree 5 with a common denominator.
The Gaussian noises $e_i$ have variance $\sigma^2=7.243$
which corresponds to $0.3$ times the variance of the process $\sum_{k=1}^{100} F_k u_k$.
Model \eqref{MM2} is adopted
with each impulse response of order $p=50$ and only results coming from RSGSOB and RSGS are shown,
letting $\alpha=0.9$, $n_{OB}=10$ and $n_{MC}=1000$.
The pie charts in Fig. \eqref{fig_ex2_prob} (b) display the frequencies
with which the blocks 
have been selected after running RSGSOB.
The one in the right panel regards all the blocks $\{b; b\in \mathcal{M}\}$.
One can see that the frequency of sampling blocks in $\mathcal{M}_2$ is close to $\sum P_{\mathcal{M}}(b \in \mathcal{M}_2)=1/11$. The middle panel
displays the frequencies concerning only the blocks $\{b; b\in \mathcal{M}_2\}$, i.e.,
this is a pie chart specific for the $8.9\%$ marked data in the right panel.
The same colorbar adopted in Fig.~\eqref{fig_ex2_prob} (a) is used, with
the index pairs as labels for different $\theta_{ij}$.
As expected, 
the pair $(i,j)$ is sampled more frequently according to the growth of the correlation
indicated by the $P_{ij}$ values.
During the simulation none of the blocks $\theta_{ij} \in \mathcal{M}_2$ with $i,j>10$ is sampled
due to very small associated $P_{ij}$ values.

\begin{table}[!t]
\centering
\caption{Hyperparameters and convergence rates in Example 2}\label{table_ex2covg}
\setlength{\tabcolsep}{2.4mm}{
\begin{tabular}{|c|c|c|c|} \hline
   \diagbox[dir=SE]{Algorithm}{Variables}& $\hat{\lambda}$ & $\hat{\sigma}^2$ & $rate$ \\ \hline
   RSGS & 0.6785 & 7.730 & 0.9930  \\ \hline
   RSGSOB & 0.6790 & 7.722 &  0.8919 \\ \hline
\end{tabular}}
\end{table}

\begin{figure*}[htbp]
\centering
\includegraphics[scale=0.38]{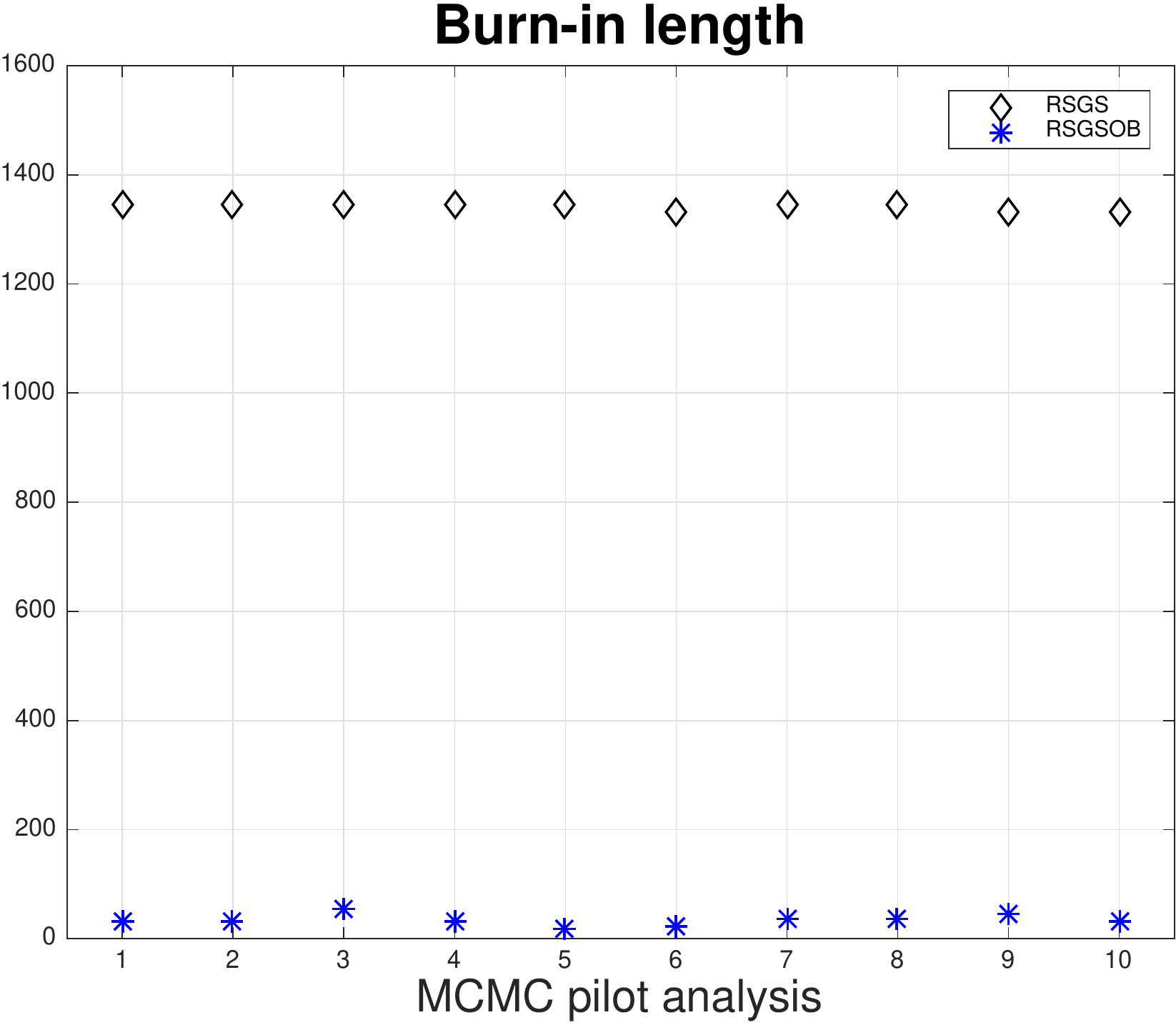} \hspace{1cm} \ \includegraphics[scale=0.38]{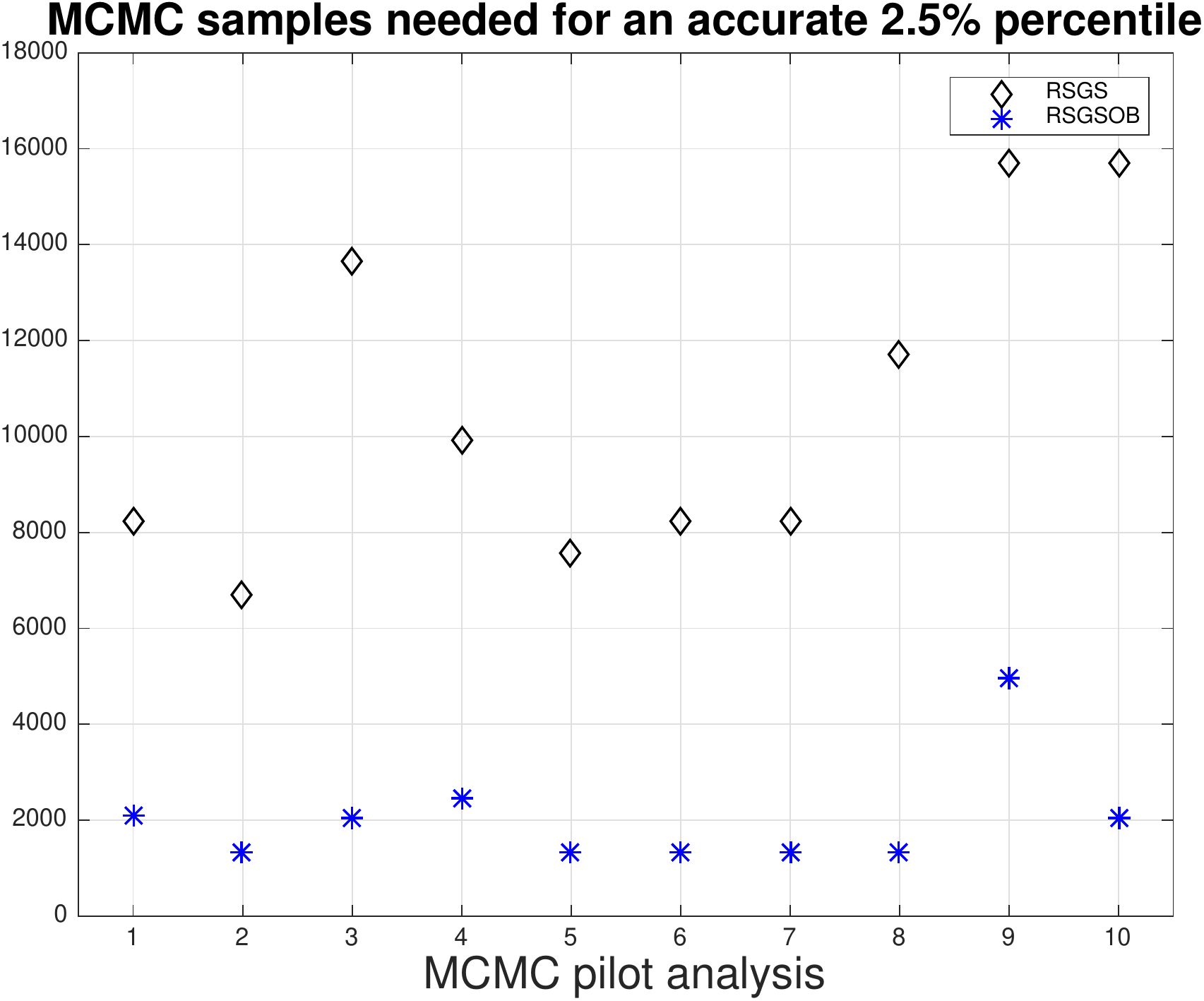}
\caption{Raftery-Lewis applied to 10 pilot analyses generated by
RSGSOB and RSGS: maximum values of the burn-in length (left) and of
the number of samples needed to estimate the percentile $2.5\%$ of the coefficients of the first 10 impulse responses
with accuracy 0.02 and probability 0.95.}
\label{FigRL}
\end{figure*}

To better appreciate the difference between the results coming from  RSGSOB and RSGS, 
first it is useful to compute their convergence rates. This is obtained using, respectively, Theorem \ref{thmcon} and Proposition \ref{thmcon2},
with hyperparameters set to their estimates $\hat{\lambda}$ and $\hat{\sigma}^2$
computed from the first 200 MCMC samples.
Results are shown in TABLE~\ref{table_ex2covg}: the convergence rates
are 0.89 for RSGSOB and 0.99 for RSGS, suggesting that
the proposed algorithm will converge much faster.
To quantify this aspect we adopt the Raftery-Lewis criterion (RLC) described in
\cite[Chapter 7]{Gilks}. Such algorithm is given a pilot analysis consisting
of the first samples generated by an MCMC scheme. Then, it determines in advance the number of
initial samples $M$ that need to be discarded to account for the burn-in and the number of samples
$N$ necessary to estimate percentiles of the posterior of any parameter with a certain accuracy.
We generate 10 pilot analyses, each consisting of the first 200 samples generated by
RSGSOB and RSGS. Then, we use RLC  to obtain the maximum values of $M$ and
$N$ to estimate the percentile $2.5\%$ of the coefficients of the first 10 impulse responses
with accuracy 0.02 and probability 0.95.
Results are displayed in Fig. \ref{FigRL}. The average of the burn-in length $M$ for
RSGSOB and and RSGS is, respectively, around 33 and 1340.
The average of the required samples $N$
is instead, respectively, around 2000 and 10500. These results well point out the computational advantages
of RSGSOB: the use of overlapping blocks much decreases the number of iterations needed
to converge to the target posterior and to visit efficiently its support. In particular,
in view of the small burn-in, samples from the desired posterior can be obtained after few MCMC steps.\\

\begin{figure*}[htbp]
\centering
\includegraphics[scale=0.56]{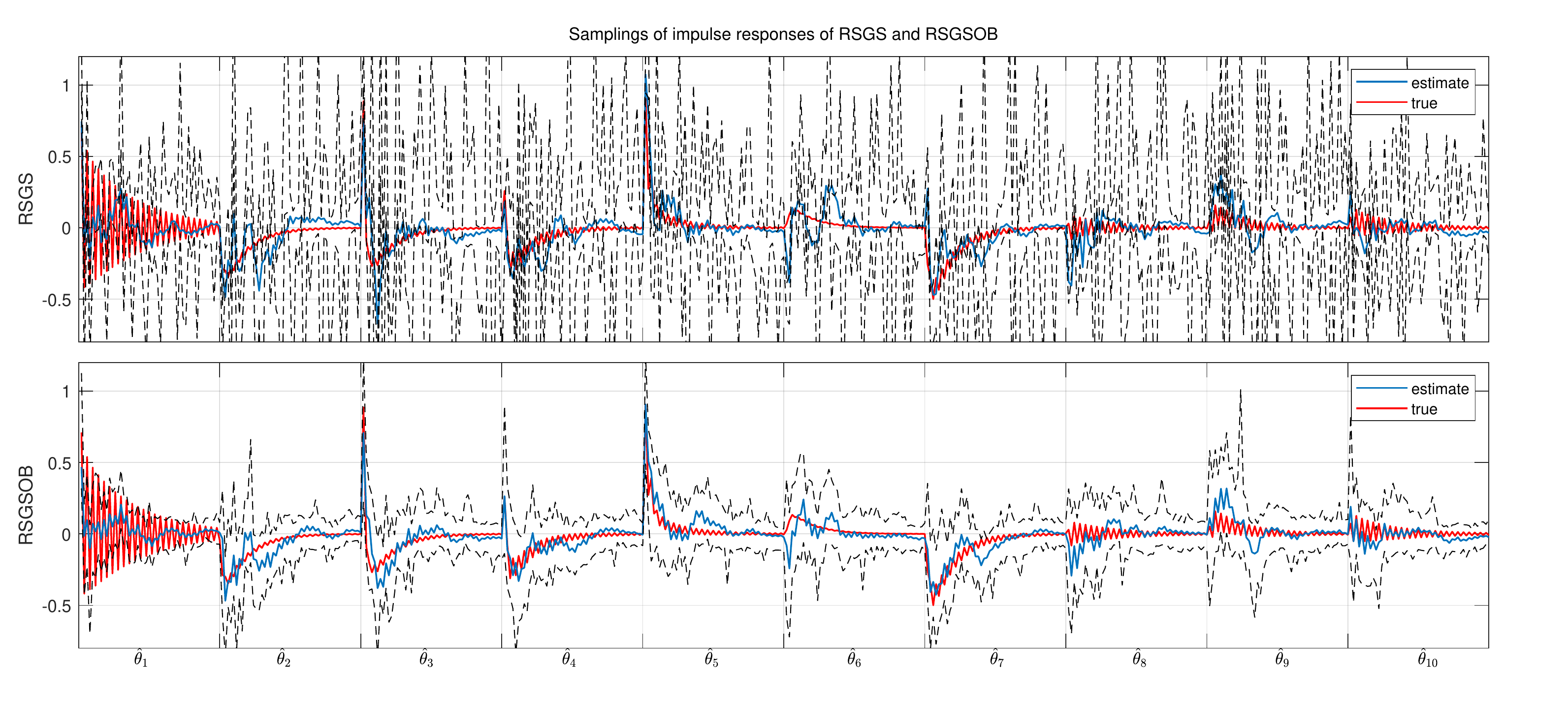}
\caption{True impulse responses (red) together with
the mean (blue) and the $95\%$ uncertainty bounds (dashed) computed by the first 100 samples of $\{\theta\}_{i=1}^{10}$.}
\label{FigMCMC100}
\end{figure*}

\begin{figure*}[htbp]
\centering
\includegraphics[scale=0.56]{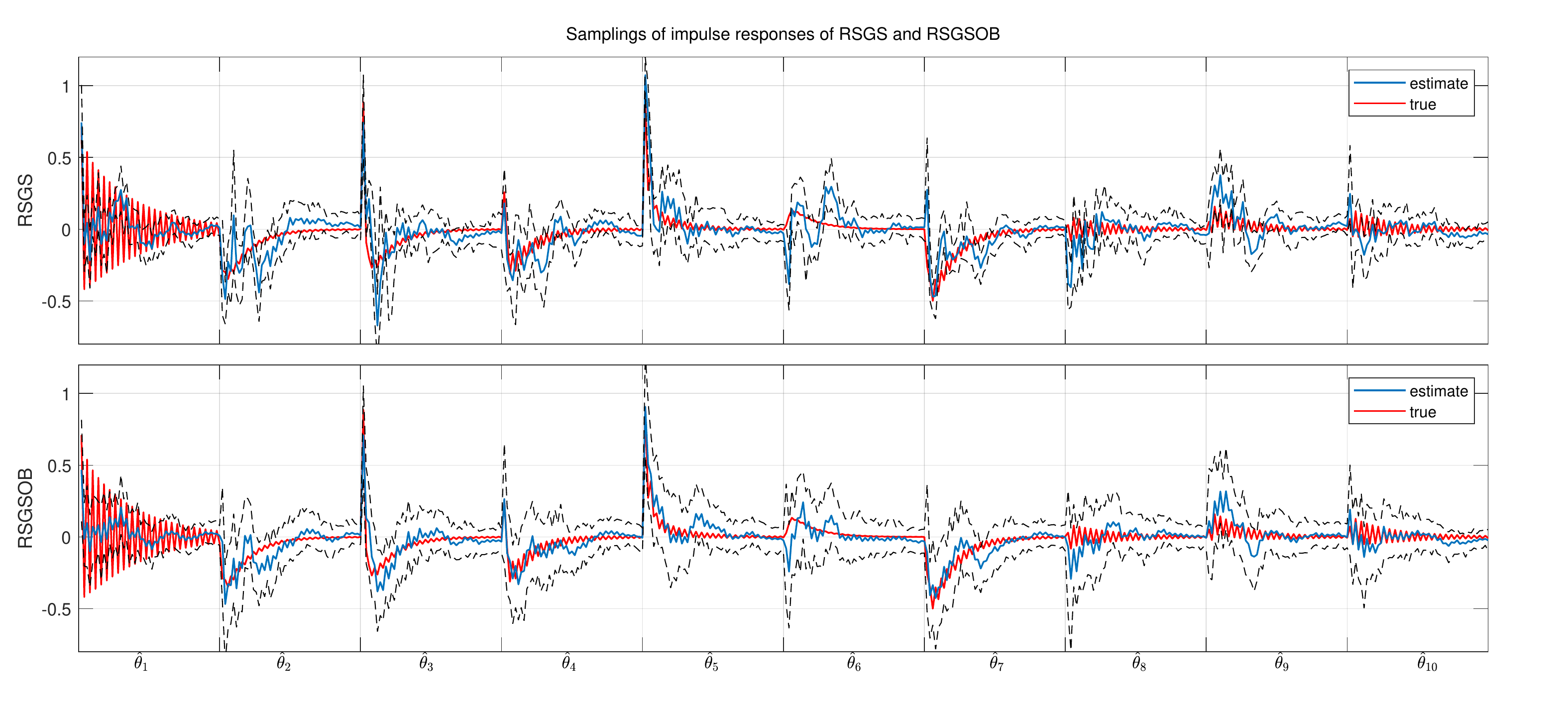}
\caption{True impulse responses (red) together with
the mean (blue) and the $95\%$ uncertainty bounds (dashed)  computed by the first 200 samples of $\{\theta\}_{i=1}^{10}$.}
\label{FigMCMC200}
\end{figure*}

Now, the MCMC estimates obtained by RSGSOB and RSGS are illustrated.
First, we focus on the first 10 impulse responses related to the strongly collinear inputs.
Fig. \ref{FigMCMC100} shows the
true impulse responses (red) together with the posterior mean (blue) and the 95$\%$ uncertainty bounds (dashed)
computed by the first 100 samples. While RSGSOB already returns good estimates and credible confidence intervals,
the outcomes from RSGS are not reliable. The same results after 200 iterations are displayed in Fig. \ref{FigMCMC200}.
RSGS results improve but are still quite far from those returned by RSGSOB.\\

The quality of the estimators is finally measured by computing the fit measures: 
given an unknown vector $x$ and its estimate $\hat{x}$, it is given by
$$
 {fit}(\hat{x}):=100\Big(1- \frac{\|x-\hat x\|}{\| x \|}\Big),
$$
with $\| \cdot \|$ to indicate the Euclidean norm. 
Specifically, the fits for algorithms RSGS and RSGSOB are compared in TABLE~\ref{table_ex2fit},
for different number of MCMC samples used to compute the posterior mean and setting $x$ to three different vectors
$\theta$, $\theta_{col}$
and  $\theta_{ind}$.
The first vector $\theta$ contains all the 100 impulse responses
while the second gathers only the first 10 impulse responses (hard to estimate due to collinearity), i.e.
$$
  \theta_{col}= [\theta_1', \cdots, \theta_{10}' ]'.
$$
Finally, the third vector contains the remaining 90 impulse responses (easier to estimate)
$$
    \theta_{ind}= [\theta_{11}', \cdots, \theta_{100}' ]'.
$$


\begin{table}[!t]
\centering
\caption{sum of fits in Example 2}\label{table_ex2fit}
\setlength{\tabcolsep}{1mm}{
\begin{tabular}{|c|c|c|c|}
  \hline
  \diagbox[dir=SE]{Algorithm}{100 \ iterations} & $fit(\hat \theta)$ & $fit(\hat{\theta}_{col})$ & $fit(\hat{\theta}_{ind})$ \\ \hline
  RSGS & -15.2 & -231.2 & 71.6 \\ \hline
  RSGSOB & 65.7 & 29.5 & 73.9\\
  \hline
\end{tabular}
\begin{tabular}{|c|c|c|c|}
  \hline
  \diagbox[dir=SE]{Algorithm}{200 \ iterations} & $fit(\hat \theta)$ & $fit(\hat{\theta}_{col})$ & $fit(\hat{\theta}_{ind})$ \\ \hline
  RSGS & 61.2 & 10.9 & 74.0\\ \hline
  RSGSOB & 65.5 & 28.5 & 73.9\\
  \hline
\end{tabular}
\begin{tabular}{|c|c|c|c|}
  \hline
  \diagbox[dir=SE]{Algorithm}{1000 iterations} & $fit(\hat \theta)$ & $fit(\hat{\theta}_{col})$ & $fit(\hat{\theta}_{ind})$ \\ \hline
  RSGS & 65.1 & 25.2 & 74.5\\ \hline
  RSGSOB & 66.4 & 30.2 & 74.6\\
  \hline
\end{tabular}
\begin{tabular}{|c|c|c|c|}
  \hline
  \diagbox[dir=SE]{Algorithm}{2000 iterations} & $fit(\hat \theta)$ & $fit(\hat{\theta}_{col})$ & $fit(\hat{\theta}_{ind})$ \\ \hline
  RSGS & 65.8 & 27.7 & 74.7\\ \hline
  RSGSOB & 66.6 & 30.9 & 74.7\\
  \hline
\end{tabular}}
\end{table}

As expected,  the performance is similar when $\theta_{ind}$ has to be estimated
while the best possible fits for $\theta_{col}$ are reached by RSGSOB after few iterations, confirming
the predictions of the RLC.

\section{Conclusion}\label{secConclu}
Identification of large-scale linear dynamic systems may be complicated by two important factors.
First, the number of unknown variables to be estimated may be large.
So, when the problem is faced in a Bayesian framework using MCMC to reconstruct the posterior
 of the impulse response coefficients in $\theta$, 
any step of the algorithm can be computationally expensive. In fact,
chain generation requires to invert a matrix of very large dimension 
to draw samples from the full conditional distributions of $\theta$.
Hence, it is important to design strategies where small groups of variables
are updated. Second, ill-conditioning can affect the problem due to input collinearity.
This problem can be e.g. encountered in the estimation of dynamic networks where feedback
and algebraic loops are often present. It leads to slow mixing of the generated Markov chains.
This aspect requires a careful selection of the blocks of variables to be sequentially updated.
An MCMC strategy for identification of large-scale dynamic systems based on the stable spline prior has been here proposed to address both of these issues.
It relies on Gibbs sampling complemented with a strategy which also updates some relevant overlapping blocks.
The updating frequencies of such blocks are regulated by the level of collinearity among different system inputs.
The proposed algorithm has been analyzed under a theoretical perspective,
proving its convergence and deriving also the convergence rate, also showing
its effectiveness through simulated studies.



\bibliographystyle{plain}        
\bibliography{biblio}

%


\appendix
\section{Proof of Theorem \ref{thmcon}}
\subsection{Preliminary lemmas}\label{apdxpdf}
We start reporting five lemmas which are instrumental for the proof of RSGSOB convergence.
The first three derive from direct calculations which exploit
the algorithmic structure of RSGSOB.  In particular, the first one
gives useful distributions in Section~\ref{secConvg}.

\begin{lem}[$p(\theta^{s}\mid  Y, \sigma^{2(t)},\lambda^{(t)}, \theta^{s-1}, b)$]\label{prop1}
At the $t$-th iteration of Algorithm 1, given a chosen block $b$, the conditional distribution of $\theta$ from sub-iteration step $s-1$ to $s$ is Guassian, i.e.,
\begin{equation}\label{thetas}
 \theta^{s}\mid Y,\sigma^{2(t)},\lambda^{(t)},\theta^{s-1}, b \sim \mathcal{N}(\hat{\mu}^{s},\hat{\Sigma}^{s}).
\end{equation}
Two cases now arises:\\
{\bf When $b=\theta_{ij}$},
\begin{subequations}\label{muSigmas_OB}
\begin{equation}
    \hat\mu^{s}=C_{ij} \theta^{s-1} + c_{ij},
\end{equation}
and, letting $\hat{\Sigma}_{ij}$ be defined in \eqref{Eij} with
$\lambda_i=\lambda_j=\lambda^{(t)}$, the conditional density \eqref{thetas} is fully specified by
\begin{equation}
 \hat{\Sigma}^{s}=\begin{bmatrix}
                     0_{i-1} &  &  &   &      \\
                       & \hat{\Sigma}_{ij}(1,1)  &  &  \hat{\Sigma}_{ij}(1,2)  &   &  \\
                       &   &  0_{j-i-1} &    &     \\
                       &  \hat{\Sigma}_{ij}(2,1)  &   &  \hat{\Sigma}_{ij}(2,2)  &    \\
                       &   &   &    &    0_{m-j}
                   \end{bmatrix},
\end{equation}
i.e., $ \hat{\Sigma}^{s}$ is a zero block matrix except for the intersections of the $i$-th and $j$-th rows and columns being the entries of matrix $\hat{\Sigma}_{ij}$, and by
\begin{equation}\label{Cij}
  C_{ij}=\left[ \begin{array}{ccccc}
           I &   &   &   &   \\
             & I &   &   &   \\
             &  & \ddots &   &   \\
           \multicolumn{5}{c}{ \text{---}~~D_{ij}^i~~\text{---}}   \\
             &  & \vdots  &   &   \\
             \multicolumn{5}{c}{ \text{---}~~D_{ij}^j~~\text{---}}  \\
               &  &   &  \ddots &   \\
             &  &   &   &  I
         \end{array}\right],
\end{equation}
\begin{equation}
\begin{split}
  c_{ij}=[0,\cdots,~&([I,~0]\hat{\Sigma}_{ij}\frac{1}{\sigma^{2(t)}}G_{ij}'Y)',\\
    & \cdots,~([0,~I]\hat{\Sigma}_{ij}\frac{1}{\sigma^{2(t)}}G_{ij}'Y)',\cdots,0]',
\end{split}
\end{equation}
\begin{equation}
  D_{ij}^i= -\begin{bmatrix}I &0 \end{bmatrix} \hat{\Sigma}_{ij}\frac{1}{\sigma^{2(t)}}G_{ij}' G_{(ij)},
\end{equation}
\begin{equation}
  D_{ij}^j= -\begin{bmatrix}0 &I \end{bmatrix} \hat{\Sigma}_{ij}\frac{1}{\sigma^{2(t)}}G_{ij}' G_{(ij)},
\end{equation}
\begin{equation}
  G_{(ij)}= \begin{bmatrix}
          G_1 & \cdots & 0 & \cdots & 0 & \cdots & G_m
      \end{bmatrix}.
\end{equation}
\end{subequations}
i.e., $C_{ij}$ is a block identity matrix except for the $i$-th row $D_{ij}^{i}$ and  the $j$-th row $D_{ij}^{j}$,
$c_{ij}$ is a sparse block vector with the $i$-th and $j$-th entries equal to the blocks of matrix $\hat{\Sigma}_{ij}\frac{1}{\sigma^{2(t)}}G_{ij}'Y$, $G_{(ij)}$ is similar to the matrix $G$ except for $i$-th and $j$-th block zero.

{\bf When $b=\theta_i$}, similarly,
\begin{subequations}\label{muSigmas}
\begin{equation}
    \hat\mu^{s}= C_i\theta^{s-1} +c_i,
\end{equation}
\begin{equation}
    \hat{\Sigma}^{s}= {\rm diag}\{0, \cdots, \hat{\Sigma}_i, \cdots, 0\},
\end{equation}
with
\begin{equation}\label{Ci}
   C_{i}=\left[ \begin{array}{ccccc}
           I &   &   &  \\
             &   \ddots &     \\
           \multicolumn{4}{c}{ \text{---}~~D_{i}~~\text{---}}   \\
               &  &   \ddots &   \\
             &  &     &  I
         \end{array}\right]
\end{equation}
\begin{equation}
 c_{i}= [0, \cdots, (\hat{\Sigma}_i\frac{1}{\sigma^{2(t)}}G_i'Y)', \cdots, 0]',
\end{equation}
\begin{equation}
 D_i= - \hat{\Sigma}_i\frac{1}{\sigma^{2(t)}}G_i'G_{(i)},
\end{equation}
\begin{equation}
  G_{(i)}=\begin{bmatrix}
          G_1 & \cdots & 0 & \cdots & G_m
      \end{bmatrix},
\end{equation}
and $\hat{\mu}_i$, $\hat{\Sigma}_i$ from \eqref{theta_k}-\eqref{Sigmak} given $\theta^{s-1}$, $\sigma^{2(t)}$ and the same scale factor $\lambda^{(t)}$.
\end{subequations}
\end{lem}

\begin{lem}[$\pi^{(t)}(\theta^{s} \mid  \theta^{s-1})$]\label{prop2}
Denote the transfer probability of $\theta$ from line $5$ to line $9$ at iteration time $t$, and from the random sampling step $s-1$ to $s$, by $\pi^{(t)}(\theta^{s}\mid  \theta^{s-1})$. Then we have
\begin{equation}\label{piss-1}
\begin{split}
    \pi^{(t)}(\theta^{s}&\mid  \theta^{s-1})
    = p(\theta^{s}\mid  Y, \sigma^{2(t)},\lambda^{(t)}, \theta^{s-1}) \\
     &= \sum_{\mathcal{M}}P_{\mathcal{M}}(b)p(\theta^{s}\mid Y, \sigma^{2(t)},\lambda^{(t)}, \theta^{s-1},b),
\end{split}
\end{equation}
where $p(\theta^{s}\mid Y,\sigma^{2(t)},\lambda^{(t)},\theta^{s-1},b)$
is given by Lemma~\ref{prop1}
and $P_{\mathcal{M}}(b)$ is given by \eqref{PM}.
\end{lem}

\begin{lem}[transition kernel]\label{propker}
Given $\theta^{(t-1)}$, $\sigma^{2(t)}$, $\lambda^{(t)}$, denote by
\begin{equation}\label{pis}
  \pi^{(t)}(\theta^{s}):= p(\theta^{s}\mid  Y, \sigma^{2(t)},\lambda^{(t)}, \theta^{(t-1)}).
\end{equation}
The transition kernel of our Algorithm~\ref{algRSGSOB}, i.e., RSGSOB, from iteration step $t-1$ to $t$ is
\begin{equation}\label{transK}
\begin{split}
    &K(\theta^{(t)}, \sigma^{2(t)},\lambda^{(t)} \mid \theta^{(t-1)}, \sigma^{2(t-1)},\lambda^{(t-1)})\\
    =&p(\theta^{(t)}\mid  Y, \sigma^{2(t)},\lambda^{(t)}, \theta^{(t-1)})
     p(\sigma^{2(t)} \mid Y, \lambda^{(t)}, \theta^{(t-1)})\\
     &~~\cdot p(\lambda^{(t)} \mid Y, \sigma^{2(t-1)}, \theta^{(t-1)}),
\end{split}
\end{equation}
where the conditional distributions $\sigma^{2(t)} \mid Y, \lambda^{(t)}, \theta^{(t-1)}$ and $\lambda^{(t)} \mid Y, \sigma^{2(t-1)}, \theta^{(t-1)}$ are two inverse gamma distributions from \eqref{gibbsLambda}, \eqref{gibbsSigma2}, and
\begin{subequations}\label{nestedInt}
\begin{equation}
\begin{split}
    p(\theta^{(t)}\mid  Y, \sigma^{2(t)},\lambda^{(t)}, \theta^{(t-1)})
     =\pi^{(t)}(\theta^{m+n_{OB}}),
\end{split}
\end{equation}
where
\begin{equation}
    \pi^{(t)}(\theta^{s})= \int_{D_\theta} \pi^{(t)}(\theta^{s}|\theta^{s-1})\pi^{(t)}(\theta^{s-1}){\rm d}\theta^{s-1}, ~s\geq2,
\end{equation}
\begin{equation}
    \pi^{(t)}(\theta^{1})=\pi^{(t)}(\theta^{1}|\theta^{0}),
\end{equation}
$\pi^{(t)}(\theta^{s}|\theta^{s-1})$ is given by Lemma~\ref{prop2}.
\end{subequations}
\end{lem}

\bigskip

The fourth lemma is an important convergence result obtained in \cite{RS97},
while the last one regards the spectral radius of a matrix power whose proof is reported for the sake of completeness.
\begin{lem}[Theorem 3, \cite{RS97}]\label{thmRob3}
Consider the following random scan sampler. Let $C_i$, $1\leq i \leq n$, be $m\times m$ matrices and $\Psi_i$, $1\leq i \leq n$, be $m\times m$ non-negative definite matrices.  Given $\theta^{(t)}$, $\theta^{(t+1)}$ is chosen from $$N(C_i\theta^{(t)}+c_i, \Psi_i),$$
with probability $p_i$, $0\leq p_i \leq 1$ for each $i$ and $\sum p_i=1$. Thus at each transition the chain chooses from a mixture of autoregressive alternatives.  For this chain, the convergence rate is the maximum modulus eigenvalue of the matrix
$$
C=\sum_{i=1}^{n}p_iC_i.
$$
\end{lem}

\begin{lem}\label{lem2}
The spectral radius of a matrix power is the power of the matrix's spectral radius, i.e.,
\begin{equation}\nonumber
 \rho(A^m)=\rho(A)^m.
\end{equation}
\end{lem}
\begin{pf}
Any matrix $A$ can be written as $A=PJP^{-1}$, with $J$ the Jordan canonical form of $A$, then $C^{m}=PJ^{m}P^{-1}$. And hence the eigenvalues of $A^m$ are the $m$ power of all eigenvalues of $A$.
Denote by $\varphi(A)$ the characteristic polynomial of $A$.
Since for any complex number $\zeta \in \mathbb{C}$, $|\zeta^m| = |\zeta|^m$, we have
\begin{equation}\nonumber
\begin{split}
  \rho(A^m) &=\max\{|\zeta|; \varphi_{A^m}(\zeta)=0\} \\
   &=\max\{|\zeta^m|; \varphi_{A}(\zeta)=0\}\\
   &= \max\{|\zeta|^m; \varphi_{A}(\zeta)=0\} =\rho(A)^m.
\end{split}
\end{equation} \hfill $\Box$
\end{pf}

\subsection{Proof of Theorem \ref{thmcon}}\label{apdxproof}

To prove the first part of the theorem regarding the convergence of RSGSOB, the first step is to show that
the Markov chain generated by this algorithm has invariant density $p(\cdot)$ given by the
the target posterior. This is a consequence of the Metropolis-Hastings algorithm.
The update of $\sigma^{2}$ and $\lambda$ done inside each step of RSGSOB is standard
since it exploits their full conditional distributions. Hence, the invariance of $p$ is guaranteed.
Now, let us focus on the update of $\theta$ and use
$p(\cdot|\cdot)$
to indicate the transition density that selects randomly a block and updates it.
Such transition density can be written as $p(\cdot|\cdot)=\sum_b
P_b p_b(\cdot|\cdot)$
where $\sum_b P_b=1$ and each $p_b(\cdot|\cdot)$ can update
either just one of the impulse responses $\theta_k$ (type one)
or the couple $\theta_k,\theta_j$ denoted by $\theta_{kj}$ (type two)
through full-conditional distributions. Without loss of generality, we
change the values of these proposal densities over a set of null
probability measure. Specifically, for type one we set  $p_b(\theta^*|\theta^\#)=0$ if
the block $\theta_k$ contained in $\theta^*$ contains the same values of the
block $\theta_k$ in $\theta^\#$.
For type two, we set  $p_b(\theta^*|\theta^\#)$ if the vector $\theta_k$ in $\theta^*$
coincides with the vector $\theta_k$ in $\theta^\#$ or if the vector
$\theta_j$ in $\theta^*$ coincides with the vector $\theta_j$ in $\theta^\#$.

Assume that two vectors $\theta^{s}$ and $\theta^{s-1}$ are compatible with the evolution of RSGSOB after one sub-iteration,
i.e. they may differ only in the values contained e.g. in the $b$-th block.
The acceptance rate is
$$
  \min \left\{~1,~\frac{p(\theta^{s})\sum_b P_b p_b(\theta^{s-1} \mid \theta^{s})}{p(\theta^{s-1})\sum_b P_b p_b(\theta^{s} \mid \theta^{s-1})}~\right\}
$$
$$
=\min \left\{~1,~\frac{p(\theta^{s}) p_{b^*}(\theta^{s-1} \mid \theta^{s})}{p(\theta^{s-1}) p_{b^*}(\theta^{s} \mid \theta^{s-1})}~\right\},
$$
where the last equality comes from the fact that
$p_b( \theta^{s} | \theta^{s-1})=p_b( \theta^{s-1} | \theta^{s})=0$ for any $b\ne b^*$.
Then, by Bayes rules,
\begin{equation}
\begin{split}
    &\frac{p(\theta^{s})}{p(\theta^{s-1})p_{b^*}(\theta^{s} \mid \theta^{s-1})}p_{b^*}(\theta^{s-1} \mid \theta^{s}) \\
    =& \frac{p(\theta^{s})}{p(\theta^{s-1})p_{b^*}(\theta^{s} \mid \theta^{s-1})} \frac{p_{b^*}(\theta^{s} \mid \theta^{s-1})p(\theta^{s-1})}{p(\theta^{s})}\\
    =&1.
\end{split}
\end{equation}
Since this holds for any block $b^*$, the acceptance rate is always equal to one.
This means that, if the selected block 
is accepted with probability 1, the Metropolis-Hastings rule is followed.
This is exactly what is done by RSGSOB and this implies that such algorithm generates 
a Markov chain with invariant density $p(\cdot)$.\\ 
Now, to complete the first part of the proof of the theorem, we need to prove that the Markov chain converges
to the invariant density.
For an MCMC algorithm, this holds if the chain is irreducible and aperiodic \cite{Gilks}.
These properties are satisfied by RSGSOB.
In fact, in \eqref{transK} the conditional distributions of $\sigma^{2(t)}$ and $\lambda^{(t)}$ are inverse gamma from \eqref{gibbsLambda}, \eqref{gibbsSigma2}, hence are continuous and greater than $0$ in the regions $D_{\sigma^2}= \mathbb{R_+}$, $D_\lambda= \mathbb{R_+}$.
Since $\pi^{(t)}(\theta^{s}\mid  \theta^{s-1})$ in \eqref{piss-1} is a weighted sum of Gaussian distributions from \eqref{piss-1}\eqref{thetas}-\eqref{muSigmas}, it is continuous and greater than zero in the region $D_\theta= \mathbb{R}^{mp}$, so as the nested integration $p(\theta^{(t)}\mid  Y, \sigma^{2(t)},\lambda^{(t)}, \theta^{(t-1)})$ from \eqref{nestedInt}.
As a result, the transition probability \eqref{transK} is continuous in $D:=D_\theta \times D_{\sigma^2} \times D_\lambda$, inducing irreducibility directly.\\
Now, letting $\gamma:= (\theta, \sigma^{2}, \lambda)$, we prove that the chain is aperiodic.
For some initial $\gamma^{(0)}$, define the region
\begin{equation}\label{Bt}
  B^{(t)} =\{\gamma^{(t)}\in D;~p(\gamma^{(t)}|\gamma^{(0)}, Y)>0\}.
\end{equation}
For any specific value $\gamma^*\in B^{(t)}$, $t\geq 1$, we have $K(\gamma^* \mid \gamma^*)>0$.
Hence there exists an open neighbourhood of $\gamma^*$, $B^{(t)*} \subseteq B^{(t)}$, and $\varepsilon(\gamma^*)>0$ such that, for all $\gamma^\# \in B^{(t)*}$,
$$K(\gamma^* \mid \gamma^\#)\geq \varepsilon(\gamma^*) >0,$$
i.e., lower semi-continuous at $0$ (see \cite{Roberts94}).
It follows that the transition probability from the initial time to the instant $t$ satisfies
\begin{equation}\nonumber
\begin{split}
  &p(\gamma^{(t+1)}= \gamma^*|\gamma^{(0)}, Y)\\
  =& \int K(\gamma^* \mid \gamma^\#) p(\gamma^{(t)}= \gamma^\#|\gamma^{(0)}, Y){\rm d}\gamma^\# \\
  \geq & \varepsilon(\gamma^*) \int_{B^{(t)*}}p(\gamma^{(t)}= \gamma^\#|\gamma^{(0)}, Y){\rm d}\gamma^\# >0.
\end{split}
\end{equation}
Thus, from the definition in \eqref{Bt}, $\gamma^* \in B^{(t+1)}$ as well, ensuring aperiodicity and hence convergence.\\
Finally, to prove the last part of the theorem, the convergence rate is derived as follows.
From Lemma~\ref{prop1}, given $\theta^{s}$, $\theta^{s+1}$ is sampled from \eqref{thetas} with probability \eqref{PM}.
Then since this process is repeated for $m+n_{OB}$ times, and by Lemma~\ref{thmRob3}, we have
$$
rate=\rho(C^{m+n_{OB}}).
$$
Then by Lemma~\ref{lem2}, \eqref{rate} holds.
\end{document}